\documentclass[10pt,journal,compsoc]{IEEEtran}
\ifCLASSOPTIONcompsoc
  \usepackage[nocompress]{cite}
\else
  \usepackage{cite}
\fi
\ifCLASSINFOpdf
\else
\fi
\usepackage{amsmath,amssymb,graphicx}
\usepackage{amsfonts}
\usepackage{url}
\usepackage{algorithm}
\usepackage{algorithmicx}
\usepackage{algpseudocode}
\usepackage{booktabs}
\usepackage{threeparttable}
\usepackage{multirow}
\usepackage{caption}
\usepackage{array}
\usepackage{color}

\usepackage{graphicx}
\usepackage{comment}
\usepackage{amsmath,amssymb} 
\usepackage{xcolor}
\usepackage{bm}

\usepackage{verbatim}
\usepackage{multirow}
\usepackage{makecell}
\usepackage{amsmath}
\usepackage{subfigure}
\usepackage{times}
\usepackage{epsfig}
\usepackage{graphicx}
\usepackage{amsmath}
\usepackage{amssymb}

\DeclareMathOperator*{\Exp}{\mathbb{E}}

\usepackage[pagebackref=true,breaklinks=true,colorlinks,bookmarks=false]{hyperref}
\usepackage{diagbox}

\newcommand{\qileft}{[\kern-0.15em[}
\newcommand{\qiLeft}{\left[\kern-0.4em\left[}
\newcommand{\qiright}{]\kern-0.15em]}
\newcommand{\qiRight}{\right]\kern-0.4em\right]}

\newcommand{\st}{{\mbox{s.t.}}}

\newcommand{\eg}{{\emph{e.g.}}}
\newcommand{\ie}{{\emph{i.e.}}}
\newcommand{\wrt}{{{w.r.t.}}}
\renewcommand{\Roman}[1]{\uppercase\expandafter{\romannumeral#1}}

\newcommand{\orange}[1]{{\color{orange}{#1}}}
\newcommand{\FLOPs}{\mbox{FLOPs}}

\renewcommand{\c}{{\bm{c}}}

\newcommand{\w}{{\bm{w}}}

\newcommand{\B}{\mathcal{B}}
\newcommand{\C}{\mathcal{C}}
\newcommand{\D}{\mathcal{D}}
\newcommand{\E}{\mathcal{E}}

\newcommand{\I}{\mathcal{I}}
\renewcommand{\L}{\mathcal{L}}
\newcommand{\N}{\mathcal{N}}
\renewcommand{\P}{\mathcal{P}}

\newcommand{\W}{\mathcal{W}}

\begin{document}
%
\title{Searching for Network Width with Bilaterally Coupled Network}
%
%
%
%

\author{Xiu Su,~\IEEEmembership{Student~Member,~IEEE,}
        Shan You,~\IEEEmembership{Member,~IEEE,}
        Jiyang Xie,~\IEEEmembership{Student~Member,~IEEE,}\\
        Fei Wang,
        Chen Qian,
        Changshui Zhang,~\IEEEmembership{Fellow,~IEEE,}
        Chang Xu,~\IEEEmembership{Member,~IEEE}
        
\IEEEcompsocitemizethanks{
\vspace{-1mm}
\IEEEcompsocthanksitem Xiu Su and Chang Xu are with the School of Computer Science, Faculty of Engineering, The University of Sydney, Australia. E-mail: xisu5992@uni.sydney.edu.au, c.xu@sydney.edu.au \protect\\
\vspace{-1mm}
\IEEEcompsocthanksitem Jiyang Xie is with the Pattern Recognition and Intelligent Systems Lab., Beijing University of Posts and Telecommunications, China. E-mail: xiejiyang2013@bupt.edu.cn \protect\\
\vspace{-1mm}
\IEEEcompsocthanksitem Shan You, Fei Wang, and Chen Qian are with the SenseTime Research Centre. E-mail: \{youshan,wangfei,qianchen\}@sensetime.com \protect\\
\vspace{-1mm}
\IEEEcompsocthanksitem Shan You and Changshui Zhang are with the Department of Automation, Tsinghua University, Institute for Artificial Intelligence, Tsinghua University (THUAI), Beijing National Research Center for Information Science and Technology (BNRist). E-mail: zcs@mail.tsinghua.edu.cn \protect\\
}
\thanks{Manuscript received April 19, 2005; revised August 26, 2015.}}

%
%

\markboth{Journal of \LaTeX\ Class Files,~Vol.~14, No.~8, August~2015}%
{Shell \MakeLowercase{\textit{et al.}}: Bare Demo of IEEEtran.cls for Computer Society Journals}
%



\IEEEtitleabstractindextext{%
\begin{abstract}
Searching for a more compact network width recently serves as an effective way of channel pruning for the deployment of convolutional neural networks (CNNs) under hardware constraints. To fulfill the searching, a one-shot supernet is usually leveraged to efficiently evaluate the performance \wrt~different network widths. However, current methods mainly follow a \textit{unilaterally augmented} (UA) principle for the evaluation of each width, which induces the training unfairness of channels in supernet. In this paper, we introduce a new supernet called Bilaterally Coupled Network (BCNet) to address this issue. In BCNet, each channel is fairly trained and responsible for the same amount of network widths, thus each network width can be evaluated more accurately. Besides, we propose to reduce the redundant search space and present the BCNetV2 as the enhanced supernet to ensure rigorous training fairness over channels.
Furthermore, we leverage a stochastic complementary strategy for training the BCNet, and propose a prior initial population sampling method to boost the performance of the evolutionary search.
We also propose the first open-source width benchmark on macro structures named Channel-Bench-Macro for the better comparison of width search algorithms.
Extensive experiments on benchmark CIFAR-10 and ImageNet datasets indicate that our method can achieve state-of-the-art or competing performance over other baseline methods. Moreover, our method turns out to further boost the performance of NAS models by refining their network widths. For example, with the same FLOPs budget, our obtained EfficientNet-B0 achieves 77.53\% Top-1 accuracy on ImageNet dataset, surpassing the performance of original setting by 0.65\%.
\end{abstract}

}

\maketitle

\IEEEdisplaynontitleabstractindextext

%
\IEEEpeerreviewmaketitle

\IEEEraisesectionheading{\section{Introduction}\label{sec:introduction}}

%
%
%
%
\IEEEPARstart{W}{hen} deploying deep convolutional neural networks (CNNs) in the real world, it is important to take different hardware budgets into consideration \cite{dc,mobilenetv2}, e.g.,  floating point operations (FLOPs), latency, memory footprint, and energy consumption. Pruning redundant channels in CNNs is a natural way to derive a compact network that can simultaneously satisfy these different hardware constraints. Typical channel pruning usually leverages a pre-trained network and implements the pruning in an end-to-end \cite{sss,slimming} or layer-by-layer \cite{cp,tang2020reborn} manner. After pruning, the structure of the pre-trained model remains unchanged, so that the pruned network is friendly to off-the-shelf deep learning frameworks and can be further boosted by other techniques, such as quantification \cite{dc} and knowledge distillation \cite{distilling}.

Recently, the core of channel pruning has been suggested to learn a more compact \textit{network width} instead of the retained weights \cite{rethinkingpruning}. The number of channels or filters is taken as a direct measure of the network width \cite{metapruning,autoslim,tas}. Recently,  neural architecture search (NAS) \cite{yang2020ista,proxylessnas} and other AutoML techniques (such as MetaPruning \cite{metapruning}, AutoSlim \cite{autoslim}, and TAS \cite{tas}) have been explored to directly search for an optimal network width. In general, a one-shot supernet is usually adopted for the evaluation of different widths. If a certain layer is of the width $c$, we need to assign $c$ channels (filters) for that layer and all the other layers follow the similar setup. All these assigned channels in the supernet thus form a sub-network from the supernet. 

\begin{figure*}[t]
	\centering
	\includegraphics[width=0.90\linewidth]{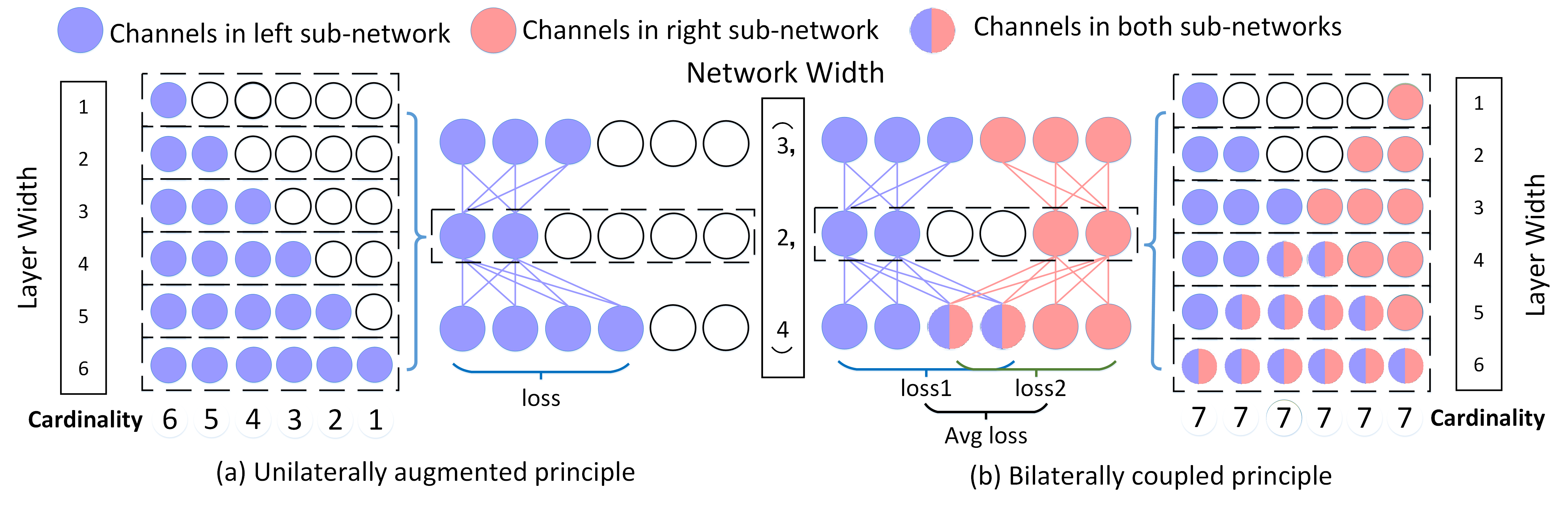}
	\vspace{-3mm}
	\caption{Comparison of unilaterally augmented (UA) principle and our proposed bilaterally coupled (BC) principle in supernet. In BC principle, each network width is indicated by two (left and right) paths, so that all channels get the same cardinality for evaluation different widths. However, in UA principle each width goes through one path, and training unfairness over channels and evaluation bias exist. Under uniform sampling strategy, for each channel the expectation of the times being evaluated is theoretically equal to the times being trained,  since we simply sample each path and train it. For simplicity,  we use \emph{cardinality} to refer to the number of times that a channel is  used for evaluation over all widths.}
	\label{motivation}
	\vspace{-4mm}
\end{figure*}

Existing methods \cite{metapruning,autoslim,tas} often follow a \textit{unilaterally augmented} (UA) principle to produce a sub-network layer of different widths from the supernet, i.e.,  channels in a layer are counted from the left to the right. To obtain a sub-network layer of a width $c$, the UA principle simply chooses the leftmost $c$ channels from the supernet. In this way, leftmost channels will be more frequently used to form the sampled sub-network of different widths, compared with the channels in the right side. For example, in Figure \ref{motivation} (a), the leftmost channel will always be used to form the sub-network layer whose width ranges from 1 to 6,  while the rightmost channel is only used in the sub-network layer that is of a width 6. This causes a \emph{training unfairness} among the channels and their corresponding kernels, i.e,  
left channels will be trained more than right ones. This training unfairness can affect the accuracy of sub-network evaluation and lead to an unconvincing optimization of the network widths.

In this paper, we introduce a new supernet called Bilaterally Coupled Network (BCNet) to address the training and evaluation unfairness within UA principle. In BCNet, each channel is fairly trained wherever it is from the left side or the right side. Specifically, both in training and evaluation, the optimality of a sub-network width is determined symmetrically by the average performance of bilateral (\ie, both left and right) channels. As shown in Figure \ref{motivation} (b), considering a supernet layer with six channels, both the leftmost channel  and the rightmost channel can always be used to form a sub-network layer whose width ranges from 1 to 6. In this way, all the channels will be trained equally and bilaterally coupled in BCNet, which leads to a more fair evaluation than that induced by the UA principle. To encourage a rigorous training fairness over channels, we adopt a complementary training strategy as shown in Figure \ref{complementary}.  

A preliminary version of this work was presented earlier \cite{su2021bcnet}, namely BCNet. This journal version adds to the initial conference paper in significant ways. First, we follow the wisdom of existing successful neural architectures from network engineering, and pre-set the smallest and maximum width allowed in a sub-network layer. The search space can be shrunk around the premise that the channel training fairness is not destroyed. Second, a new iterative updating method is developed to save half of the memory cost of BCNetV2. Third, to promote a better comparison between our algorithm with other search methods, we open-source the first width benchmark on macro structures with CiFAR-10 dataset, named Channel-Bench-Macro\footnote{https://github.com/xiusu/Channel-Bench-Macro}. The Channel-Bench-Macro contains two base models of MobileNet and ResNet, and both have 16,384 architectures and their test accuracies, parameter numbers, and FLOPs on CIFAR-10. In addition, we include more comparison experiments to demonstrate the effectiveness and advantages of the proposed algorithm. 

Extensive experiments on the benchmark CIFAR-10 and ImageNet datasets show that our method outperforms the state-of-the-art methods under various FLOPs budget. For example, our searched EfficientNet-B0 achieves 75.2\% Top-1 accuracy on ImageNet dataset with 192M FLOPs (2 $\times$ acceleration). Our method is easy to implement, and experiments prove that VGGNet \cite{vgg}, MobileNetV2 \cite{mobilenetv2}, ResNet50 \cite{dr} and even NAS-based EfficientNet-B0 \cite{efficientnet} and ProxylessNAS \cite{proxylessnas} can be further boosted using our BCNet on both the CIFAR-10 and ImageNet datasets.


    
	
	
	

\section{Related Work}

Channel pruning is an effective method to compress and accelerate an over-parameterized convolutional neural network, and thus enables the pruned network to accommodate various hardware computational budgets. Extensive studies are illustrated in the comprehensive survey \cite{survey}. Here, we summarize the typical approaches of channel pruning \cite{sss,slimming,cp,tang2020scop} and network width search methods \cite{metapruning,autoslim,tas}.

\textbf{Channel pruning.} Channel pruning is an prevalent method which aims to reduce redundant channels of an heavy model, and generally implemented by selecting significant channels \cite{slimming,cp} or adding additional data-driven sparsity \cite{sss,dcp,tang2019bringing}. For example, CP \cite{cp} propose to construct a group Lasso to select unimportant channels. Slimming \cite{slimming} impose a $\l_1$ regularization on the scaling factors. DCP \cite{dcp} propose to construct an additional discrimination-aware losses. Despite the achievements, these methods rely heavily on manually assigned pruning ratios or hyperparameter coefficients, which is complicated, time consuming and hardly to find Oracle solutions.

\textbf{Network width search.} Inspired by the development of NAS \cite{proxylessnas,guo2020hit,huang2020explicitly}, network width search methods \cite{metapruning,autoslim,tas,dmcp,amc,su2021locally} generally take a carefully designed one-shot supernet to rank the relative  performance of different widths. For example, TAS \cite{tas} aims to search the optimal network width via a learnable continuous parameter distribution.  MetaPruning \cite{metapruning} proposes to directly generate representative weights for different widths. AutoSlim \cite{autoslim} proposes to leverage a slimmable network to approximate the accuracy of different network widths.  However, all these methods follow UA principle in assigning channels, which affects the fairness in evaluation. To accurately rank the performance of network widths, our proposed BCNet aims to assign the same opportunity for channels during training, thus ensures the evaluation fairness in searching optimal widths.

\section{Channel Pruning as Network Width Search}

Formally, suppose the target network to be pruned $\N$ has $L$ layers, and each layer has $l_i$ channels. Then channel pruning aims to identify redundant channels (indexed by $\I_{pruned}^i$) layer-wisely, \ie,
\begin{equation}
\I_{pruned}^i \subset [1:l_i],
\end{equation}
where $[1:l_i]$ is an index set for all integers in the range of $1$ to $l_i$ for $i$-th layer. However, \cite{rethinkingpruning} empirically finds that the absolute set of pruned channels $\I_{pruned}^i$ and their weights are not really necessary for the performance of pruned network, but the obtained width $c_i$ actually matters, \ie,
\begin{equation}
c_i = l_i - |\I_{pruned}^i|.
\end{equation}

In this way, it is intuitive to directly search for the optimal network width to meet the given budgets. 

Denote an arbritary network width as $\c = (c_1,c_2,...,c_L) \in \C = \bigotimes_{i=1}^L [1:l_i]$, where $\bigotimes$ is the Cartesian product. Then the size of search space $\C$ amounts to $|\C| = \prod_{i=1}^{L}l_i$. However, this search space is fairly huge, \eg,  $10^{25}$ for $L=25$ layers and $l_i=10$ channels. To reduce the search space, current methods tend to search on a group level instead of channel-wise level. In specific, all channels at a layer is partitioned evenly into $K$ groups, then we only need to consider $K$ cases; there are just $(l_i/K)\cdot[1:K]$ layer widths for $i$-th layer. Therefore, the search space $\C$ is shrunk into $\C_K$ with size $|\C_K| = K^L$. In the following, we use both $\C$ and $\C_K$ seamlessly. 

During searching, the target network is usually leveraged as a supernet $\N$, and different network widths $\c$ can be directly evaluated by sharing the same weights with the supernet. 
Then the width searching can be divided into two steps, i.e., supernet training, and searching with supernet. Usually, the original training dataset is split into two datasets, \ie, training dataset $\D_{tr}$ and validation dataset $\D_{val}$. The weights $\W$ of the target supernet $\N$ is trained by uniformly sampling a width $c$ and optimizing its corresponding sub-network with weights with weights $w_c \subset W$,

\begin{equation}
W^* = \mathop{\arg\min}_{\w_\c \subset W}~ \Exp_{\c\in U(\C)} \qiLeft \L_{tr}(\w_c; \N, \c, \D_{tr})\qiRight, 
\label{eq2}
\end{equation}
where $\L_{train}$ is the training loss function, $U(\C)$ is a uniform distribution of network widths, 
and $\Exp\qiLeft\cdot\qiRight$ is the expectation
of random variables. Then the optimal network width $\c^*$ corresponds to the one with best performance on validation dataset, \eg~classification accuracy,
\begin{equation}
\begin{aligned}
\c^* = &\mathop{\arg\max}_{\c \in \C}~\mbox{Accuracy}(\c, \w^*_\c; W^*, \N, \D_{val}), \\ &~\st~\FLOPs(\c) \leq F_b,
\end{aligned}
\label{eq3}
\end{equation}	
where $F_b$ is a specified budget of FLOPs. Here we consider FLOPs rather than latency as the hardware
constraint since we are not targeting any specific hardware device like EfficientNet \cite{efficientnet} and other pruning baselines \cite{dcp,tas,amc}. The searching of Eq.\eqref{eq3} can be fulfilled efficiently by various algorithms, such as random or evolutionary search \cite{metapruning}. Afterwards, the performance of the searched optimal width $\c^*$ is analyzed by training from scratch. 

\section{BCNet: Bilaterally Coupled Network}

\begin{figure*}[t]
	\centering
	\includegraphics[width=0.82\linewidth]{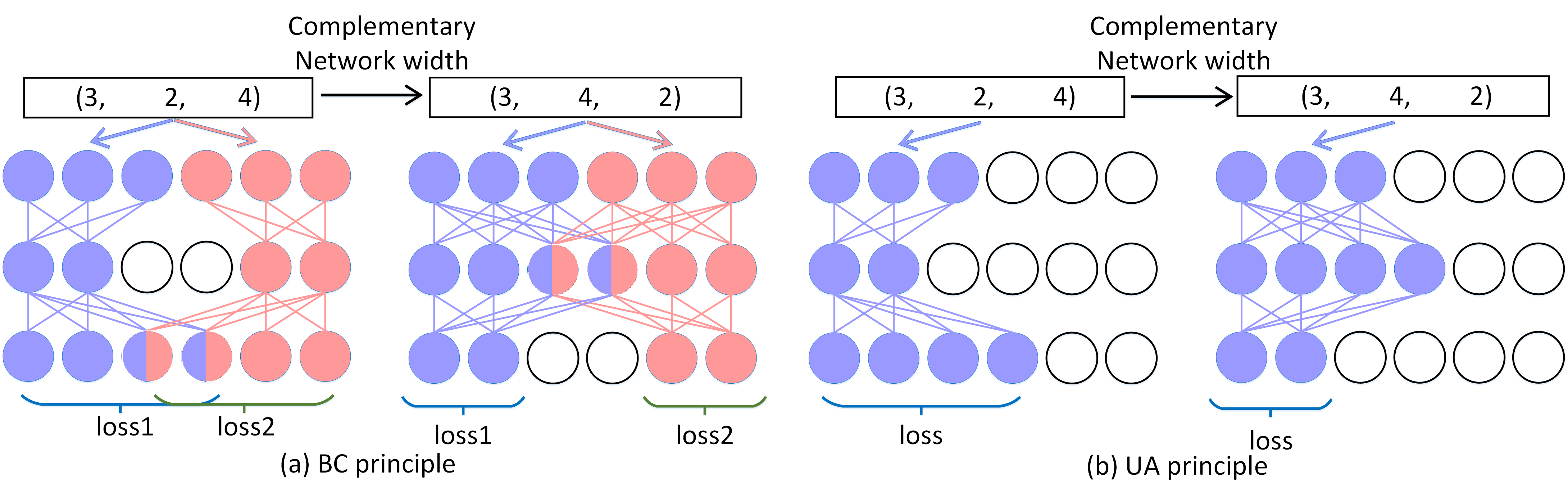}
	\vspace{-3mm}
	\caption{Illustration of the complementary of a network width for both our bilaterally coupled (BC) principle and the baseline unilaterally augmented (UA) principle. In BC principle, for any width $\c$, all channels will be trained evenly (2 times) by training $\c$ and its complementary together. However, this fairness cannot be ensured in UA principle, but gets worse; some channels will be trained 2 times while others will be trained one or zero time.}
	\label{complementary}
	\vspace{-4mm}
\end{figure*}

\subsection{BCNet as a new supernet} \label{BCNet_supernet}

As illustrated previously, for evaluation of width $c$ at certain layer, unilaterally augmented (UA) principle assigns the leftmost $c$ channels to indicate its performance as Figure \ref{motivation}(a). Hence all channels used for width $c$ can be indexed by a set $\I_{UA}(c)$, \ie,
\begin{equation}
\I_{UA}(c) = [1:c]. 
\label{ua_channel}
\end{equation}
However, UA principle imposes an unfairness in updating channels (filters) for supernet. Channels with small index will be assigned to both small and large widths. Since different widths are sampled uniformly during the training of supernet, kernels for channels with smaller index thus get more training accordingly. To quantify this training unfairness, we can use the number of times that a channel is used for the evaluation of all widths to reflect its \textit{training degree}, and we name it as  \emph{cardinality}. Suppose a layer has maximum $l$ channels, then the cardinality for the $c$-th channel in UA principle is
\begin{equation}
\mbox{Card-UA}(c) = l-c+1.
\label{ua_counts}
\end{equation}
In this way, the cardinality of all channels varies significantly and thus they get trained much differently, which introduces evaluation bias when we use the trained supernet to rank the performance over all widths.

To alleviate the evaluation bias over widths, our proposed BCNet serves as a new supernet which promote the fairness \wrt~channels. As shown in Figure \ref{motivation}(b), in BCNet each width is simultaneously evaluated by the sub-networks corresponding to left and right channels. It can be seen as two identical networks $\N_l$ and $\N_r$ bilaterally coupled with each other, and use UA principle for evaluation but in a reversed order of counting channels. In this way, all channels  $\I_{BC}(c)$ used for evaluating width $c$ in BCNet are indexed by
\begin{align}
\I_{BC}(c) &= \I_{UA}^l(c) \uplus \I_{UA}^r(c) \\
&= [1:c] \uplus [(l-c+1):l],   
\label{bc_channel}
\end{align}
where $\uplus$ means the merge of two lists with repeatable elements. In detail, left channels in $\N_l$ follow the same setting with UA principle as Eq.\eqref{ua_channel}, while for right channels in $\N_r$, we count channels starting from right with $\I_{UA}^r(c) = [(l-c+1):(l-c)]$. 
As a result, the cardinality of each channel within BC principle is the sum from both two supernets $\N_l$ and $\N_r$. In detail, since channels count from the right side within $\N_r$, the cardinality for the $c$-th channel in left side corresponds to the cardinality of $l - c + 1$-th channel in right side with Eq.\eqref{ua_counts}. As a result, the cardinality for the $c$-th channel in BC principle is 
\begin{multline}
\mbox{Card-BC}(c)  = \mbox{Card-UA}(c) + \mbox{Card-UA}(l+1-c) \\
= (l-c+1) + (l+1-l-1+c) = l+1 
\label{bc_counts}
\end{multline}
Therefore, the cardinality for each channel will always amounts to the same constant value (\ie, $7$ in Figure \ref{motivation}(b)) of widths, and irrelevant with the index of channels with BC principle, thus ensuring the fairness in terms of channel (filter) level, which promotes to fairly rank network widths with our BCNet.

\subsection{Stochastic complementary training strategy} \label{stochastic_training}
To train the BCNet, we adopt stochastic training, \ie, uniformly sampling a network width $\c$ from the search space $\C_K$, and training its corresponding channel (filters) $\N(W,\c)$ using training data $\D_{tr}$ afterwards. Note that a single $\c$ has two paths in BCNet, during training, a training batch $\B\subset\D_{tr}$ is supposed to forward simultaneously through both $\mathcal{N}^*_l(W)$ and $\mathcal{N}^*_r(W)$. Then the training loss is the averaged loss of both paths, \ie, for each batch $\B$
\begin{equation}
\begin{aligned}
\L_{tr}(W,\c;\B) = \frac{1}{2} \cdot\left(
\L_{tr}(\mathcal{N}_l; \c, \B) + \L_{tr}(\mathcal{N}_r; \c, \B)\right).
\end{aligned}
\label{eq7}
\end{equation}	

Despite with our BCNet, channels are trained more evenly than other methods. However, it still cannot ensure a rigorous fairness over channels. For example, if a layer has 3 channels and we sample 10 widths on this layer. Then results can come to that the first channel is sampled 4 times and the other two are sampled 3 times, respectively. The first channel thus still gets more training than the others, which ruins the training fairness.

To solve this issue, we propose to leverage a complementary training strategy, \ie, after sampling a network width $\c$, both $\c$ and its complementary $\bar{\c}$ get trained. For example, suppose a width $\c=(3,2,4)$ with maximum 6 channels per layer, then its complementary amounts to $\bar{\c} = (3,4,2)$ as Figure \ref{complementary}. The training loss for the BCNet is thus
\begin{equation}
\L_{tr}(W;\D_{tr},\N) = \Exp_{\bm{c}\in U(\C)} \qiLeft \mathcal{L}_{tr}(W,\bm{c};\D_{tr})  + \mathcal{L}_{tr}(W,\bar{\bm{c}};\D_{tr})\qiRight. 
\label{eq8}
\end{equation}	
In this way, when we sample a width $\c$, we can always ensure all channels are evenly trained, and expect a more fair comparison over all widths based on the trained BCNet. Note that this complementary strategy only works for our BCNet, and fails in the typical unilateral augmented (UA) principle \cite{autoslim,metapruning,tas}, which even worsens the bias as shown in Figure \ref{complementary} (b). 
\begin{figure*}[!t]
    \centering
    \includegraphics[width=0.8\linewidth]{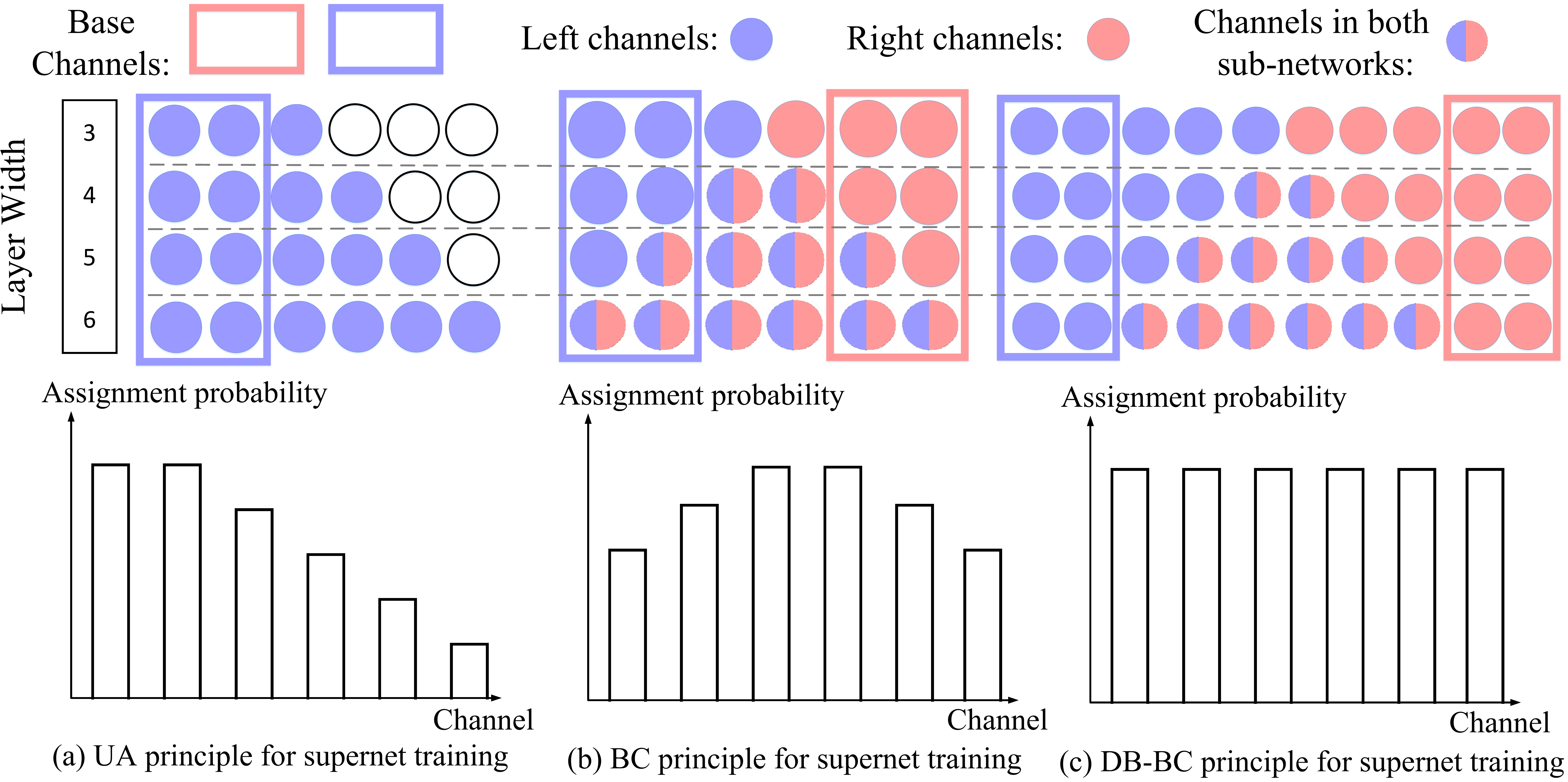}
    \vspace{-3mm}
    \caption{Comparison between (a) UA principle, (b) BC principle and (c) BCv2 principle with a top example of two base channels. In (a), channels cannot be fairly trained. Due to the appearance of the base channels (\emph{i.e.}, the channels of the base model), channels in (b) cannot obtain similar assignment probabilities. Then in (c), the proposed BCv2 overcomes the affect of the base channels and conducts a uniform assignment distribution for width.}\label{fig:DB_width}
    \vspace{-5mm}
\end{figure*}
\subsection{BCNet with evolutionary search}
After the BCNet $\N^*$ is trained with weights $W^*$, we can evaluate each width by examining its performance (\eg, classification accuracy) on the validation dataset $\D_{val}$ as Eq. \eqref{eq3}. Besides, similar to the training of BCNet, the performance of a width $\c$ is indicated by the averaged accuracy of its left and right paths, which can be formulated as Eq. \eqref{supp_evolution}. The optimal width refers to the one with the highest performance and we then train it from scratch.

Concretely, we adopt the multi-objective NSGA-\Roman{2} \cite{deb2002fast} algorithm to implement the search, and hard FLOPs constraint can be thus well integrated. Note that some networks (\eg, MobileNetV2) may have batch normalization (BN) layers, and due to the varying network widths, the mean and variance in the BN layers are not suitable to all widths. In this way, we simply use the mean and variance in batches instead, and we set the batch size to 2048 during testing to ensure accurate estimates of the mean and variance.
\begin{equation}
\begin{aligned}
&\mbox{Accuracy}(W,\c;\D_{val}) = \\
&\frac{1}{2} \cdot\left(\mbox{Accuracy}(\mathcal{N}_l, \c; \D_{val}) + 
\mbox{Accuracy}(\mathcal{N}_r, \c; \D_{val})\right).
\end{aligned}
\label{supp_evolution}
\end{equation}	
In detail, we set the population size as 40 and the maximum iteration as 50. Firstly, we apply our proposed prior initial population sampling method Eq. \eqref{eq10} $\sim$ Eq.\eqref{eq12} to generate the initial population. In each iteration,  we use the trained BCNet to evaluate each width and rank all widths in the population. After the ranking, we use the tournament selection algorithm to select 10 widths retained in each generation. And the population for the next iteration is generated by two-point crossover and polynomial mutation. Finally, the searched width refers to the one with the best performance in the last iteration, and we train it from scratch for evaluation. 


\subsection{Evolutionary search with prior initial population sampling}


In generally, evolutionary search is prone to the initial population before the sequential mutation and crossover process. In this way, we propose a Prior Initial Population Sampling method to allocate a promising initial population, which is expected to contribute to the evolutionary searching performance.

Concretely, suppose the population size is $P$, and we hope the sampled initial population have high performance in order to generate competing generations during search. Note that during training of BCNet, we have also sampled various widths, whose quality can be reflected by the training loss. In this way, we can record the top $m$ (\eg, $m=100$) widths $\{\c^{(i)}\}_{i=1}^m$ with smallest training loss $\{\ell^{(i)}\}_{i=1}^m$ as priors for good network widths. However, even the group size for every layer is set to 10, the search space of MobileNetV2 is as large as $10^{25}$, which is too large to search good widths within limited training epochs. Thus we aim to learn layer-wise discrete sampling distributions $\P(l,c_i)$ to perform stochastic sampling a width $\c = (c_1,.,c_l,.,c_L)$, where $\P(l, c_i)$ indicates the probability of sampling width $c_l$ at $l$-th layer subject to $\sum_{i}\P(l, c_i) = 1$.

Note that these $m$ prior network widths actually can reflect the preference over some widths for each layer. For example, if at a layer $l$,  a width $c_l$ exists in these $m$ prior widths with smaller training loss, then the sampling probability $\P(l,c_i)$ should be large as well. In this way, we can measure the \textit{potential error} $\E(l,c_i)$ of sampling $c_l$ width at $l$-th layer by recording the averaged training loss of all $m$ widths going through it,\ie, 
\begin{equation}
\E(l,c_i) = \frac{1}{\sum_{j=1}^m \mathbf{1}\{\c^{(j)}_l = i\}} \cdot \sum_{j=1}^m \ell^{(j)} \cdot \mathbf{1}\{\c^{(j)}_l = i\},
\label{eq10}
\end{equation}	
where $\mathbf{1}\{\cdot\}$ is the indicator function. Then the objective is to sample with minimum expected potential errors, \ie,
\begin{equation}
\begin{aligned}
\mathop{\min}_{\P}~\sum_{l}\sum_{i}&\P(l, c_i) \cdot \E(l, c_i),~\st~\sum_{i}\P(l, c_i) = 1, \\
&\P(l, c_i)\geq 0, \forall~l = 1,...,L.
\end{aligned}
\label{eq11}
\end{equation}	

\begin{figure}[t]
	\centering
	\includegraphics[width=0.7\linewidth,height=0.45\linewidth]{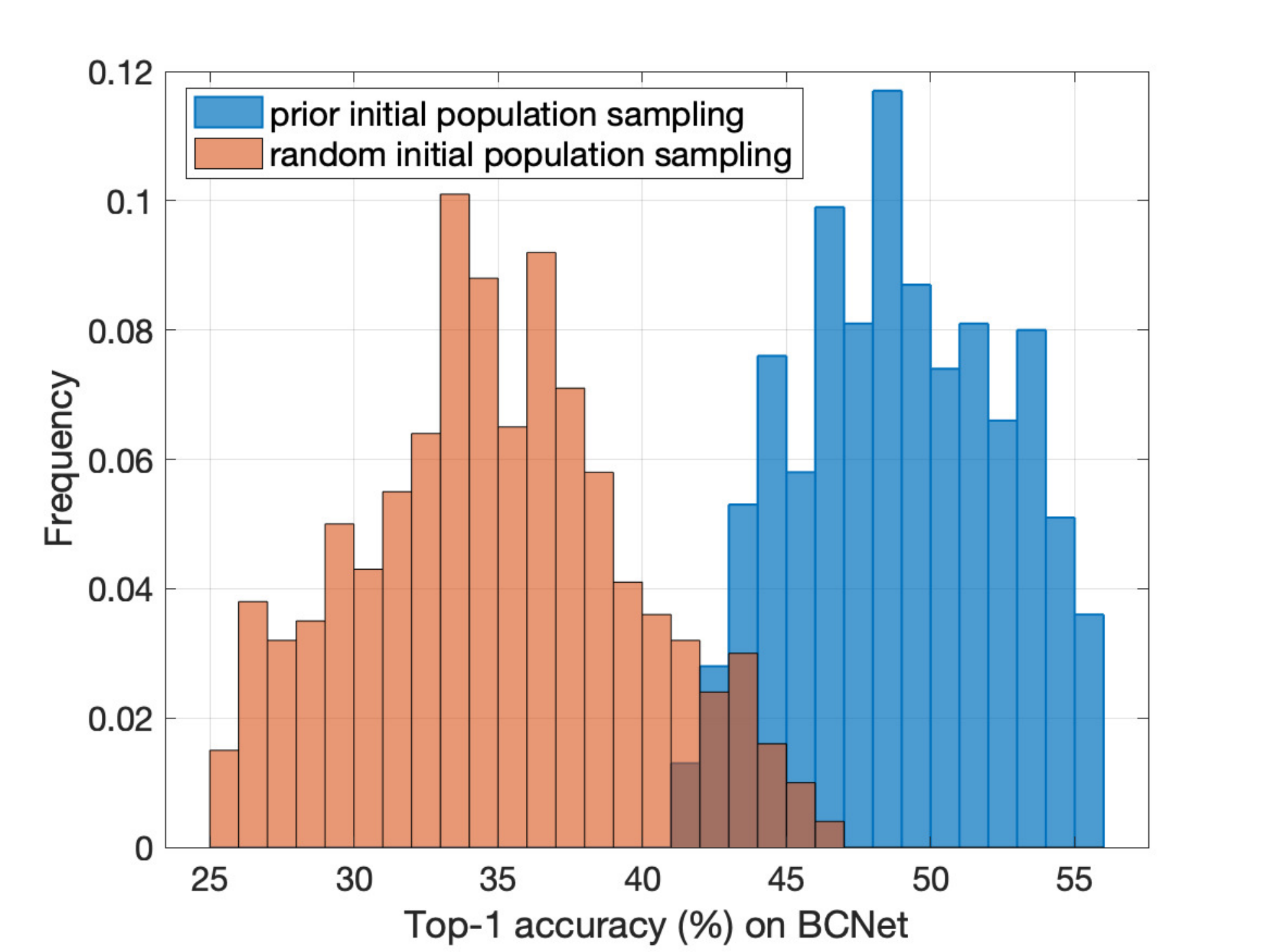}
	\caption{Histogram of Top-1 accuracy of searched widths on BCNet by evolutionary searching method with our prior or random initial population sampling \wrt~ResNet50 (2G FLOPs) on ImageNet dataset.}
	\label{prior}
	\vspace{-6mm}
\end{figure}

In addition, we also need to deal with the hard FLOPs constraint in the initial population. Since the FLOPs of a layer depends on the channels of its input and output, we can limit the expected FLOPs of the sampled network width, \ie, 
\begin{equation}
\sum_{l}\sum_{(i,j)} \P(l, c_i) \cdot F(l,c_i,c_j) \cdot \P(l+1, c_j) \leq F_b,
\label{eq12}
\end{equation}	
where $F(l,c_i,c_j)$ is the FLOPs of $l$-th layer with $c_i$ input channels and $c_j$ output channels, which can be pre-calculated and stored in a looking-up table. Then Eq.\eqref{eq12} is integrated as an additional constraint for the problem Eq.\eqref{eq11}. The overall problem is a  QCQP (Quadratically constrainted quadratic programming), which can be efficiently solved by many off-the-shelf solvers, such as CVXPY \cite{cvxpy}, GH \cite{qcqp}. As Figure \ref{prior} shows, our proposed sampling method can significantly boost evolutionary search by providing better initial populations. On average, the performance of our searched widths are much better than those obtained by random initial population, which proves the effectiveness of our proposed sampling method.

\begin{table*}[!t]
	\centering
	\scriptsize
	\caption{Performance comparison of ResNet50 and MobileNetV2 on ImageNet. Methods with "*" denotes  that the results are reported with knowledge distillation.}
	\label{Experiments_Imagenet}
	\vspace{-3mm}
	{\begin{tabular}{c|c|cc|cc||c|c|cc|cc}
			\hline
			\multicolumn{6}{c||}{ResNet50} & \multicolumn{6}{c}{MobileNetV2}\\ \hline
			FLOPs level&Methods&FLOPs&Parameters&Top-1&Top-5&FLOPs level&Methods&FLOPs&Parameters&Top-1&Top-5  \\ \hline
			\multirow{7}*{3G} & AutoSlim* \cite{autoslim} & 3.0G & 23.1M & 76.0\% & - &  \multirow{5}*{305M (1.5$\times$)} & AutoSlim* \cite{autoslim} & 305M & 5.7M & 74.2\% & - \\
			& MetaPruning \cite{metapruning} & 3.0G & - & 76.2\% & - && Uniform & 305M & 3.6M & 72.7\% & 90.7\% \\
			& LEGR \cite{legr} & 3.0G & -  & 76.2\% & - && Random & 305M & - & 71.8\% & 90.2\% \\
			& Uniform & 3.0G & 19.1M & 75.9\% & 93.0\% && \textbf{BCNet} & 305M & 4.8M & \textbf{73.9\%} & 91.5\% \\ 
			& Random & 3.0G & - & 75.2\% & 92.5\% && \textbf{BCNet*} & 305M & 4.8M & \textbf{74.7\%} & 92.2\% \\ \cline{7-12}
			& \textbf{BCNet} & 3.0G & 22.6M & \textbf{77.3\%} & 93.7\% &\multirow{12}*{200M} & Uniform & 217M & 2.7M & 71.6\% & 89.9\% \\ \cline{1-6}
			& SSS \cite{sss} & 2.8G & - & 74.2\% & 91.9\% && Random & 217M & - & 71.1\% & 89.6\% \\
			\multirow{12}*{2G} & GBN \cite{gbn} & 2.4G & 31.83M & 76.2\% & 92.8\% && \textbf{BCNet} & 217M & 3.0M & \textbf{72.5\%} & 90.6\% \\
			& SFP \cite{sfp} & 2.4G & - & 74.6\% & 92.1\% && \textbf{BCNet*} & 217M & 3.0M & \textbf{73.5\%} & 91.3\% \\
			& LEGR \cite{legr} & 2.4G & - & 75.7\% & 92.7\% && MetaPruning \cite{metapruning} & 217M & - & 71.2\% & - \\
			& FPGM \cite{fpgm} & 2.4G & - & 75.6\% & 92.6\% && LEGR \cite{legr} & 210M & - & 71.4\% & - \\
			& TAS* \cite{tas} & 2.3G & - & 76.2\% & 93.1\% && AMC \cite{amc} & 211M & 2.3M & 70.8\% & - \\
			& DMCP \cite{dmcp} & 2.2G & - & 76.2\% & - && AutoSlim* \cite{autoslim} & 207M & 4.1M & 73.0\% & - \\
			& MetaPruning \cite{metapruning} & 2.0G & -
			& 75.4\% & - && Uniform & 207M & 2.7M & 71.2\% & 89.6\% \\
			& AutoSlim* \cite{autoslim} & 2.0G & 20.6M & 75.6\% & - && Random & 207M & - & 70.5\% & 89.2\% \\
			& Uniform & 2.0G & 13.3M & 75.1\% & 92.7\% && \textbf{BCNet} & 207M & 3.1M & \textbf{72.3\%} & 90.4\% \\
			& Random & 2.0G & - & 74.6\% & 92.2\% && \textbf{BCNet*} & 207M & 3.1M & \textbf{73.4\%} & 91.2\% \\ \cline{7-12}
			& \textbf{BCNet} & 2.0G & 18.4M & \textbf{76.9\%} & 93.3\% &\multirow{6}*{150M} & AMC* \cite{amc} & 150M & - & 70.8\% & - \\ \cline{1-6}
			\multirow{11}*{1G} & AutoPruner \cite{autopruner} & 1.4G & -  & 73.1\% & 91.3\% && LEGR \cite{legr} & 150M & - & 69.4\% & - \\
			& MetaPruning \cite{metapruning} & 1.0G & - & 73.4\% & - && Uniform & 150M & 2.0M & 69.3\% & 88.9\% \\
			& AutoSlim* \cite{autoslim} & 1.0G & 13.3M & 74.0\% & - && Random & 150M & - & 68.8\% & 88.7\% \\ 
			& Uniform & 1.0G & 6.9M & 73.1\% & 91.8\% && \textbf{BCNet} & 150M & 2.9M & \textbf{70.2\%} & 89.2\% \\
			& Random & 1.0G & - & 72.2\% & 91.4\% && \textbf{BCNet*} & 150M & 2.9M & \textbf{71.2\%} & 89.6\% \\ \cline{7-12} 
			& \textbf{BCNet} & 1.0G & 12M & \textbf{75.2\%} & 92.6\% & \multirow{5}*{50M} & MetaPruning \cite{metapruning} & 43M & - & 58.3\% & - \\ 
			& AutoSlim* \cite{autoslim} & 570M & 7.4M & 72.2\% & - && Uniform  & 50M & 0.9M & 59.7\% & 82.0\%  \\ 
			& Uniform & 570M & 6.9M & 71.6\% & 90.6\% && Random & 50M & - & 57.4\% & 81.2\% \\
			& Random & 570M & - & 69.4\% & 90.3\% && \textbf{BCNet} & 50M & 1.6M & \textbf{62.7\%}& 83.7\% \\
			& \textbf{BCNet} & 570M & 12.0M & \textbf{73.2\%} & 91.1\% && \textbf{BCNet*} & 50M & 1.6M & \textbf{63.8\%}& 84.6\% \\ \hline

	\end{tabular}}	
	\vspace{-7mm}
\end{table*}

\section{Boosting the search with BCNetV2} \label{DB_BCNet}
\subsection{BCNetV2 as a supernet}

Actually, a uniform grouping method imposes a strong constraint for the partition of the search space. In specific, all channels at a layer are partitioned evenly into $K$ groups, then we only need to consider $K$ cases, where there are just $(l_i/K)\cdot[1:K]$ layer widths for the $i$-th layer. Nevertheless, searching network widths are usually implemented based on some famous pre-defined models, for example, MobileNetV2, ResNet50 and EfficientNet-B0. These architectures were carefully designed as masterpieces with countless attempts. Therefore, given a pre-set FLOPs budget, the network width search is a fine-grained and careful adjustment of the pre-defined architectures rather than disruptive redesigning of them.

With the above setting, extremely small grouping way (\eg, $l/20$) is nearly impossible to be adopted in the searched optimal network widths. Introducing inappropriate uniform groupings will increase the search space size and hamper the search for efficient network widths. Therefore, we can shrink the width range in the search by setting a smallest width $l_s$ (so-called base width) and searching between $l_s$ and maximum width $l$ to decrease the grouping number $K$. 
However, with redundant groupings removed from the search space, BC principle cannot ensure the fairness in the evaluation~\wrt~different widths. As shown in Figure \ref{fig:DB_width}, with the same setting following Section \ref{BCNet_supernet}, the assignment probability of base channels in the original BCNet is smaller than others, which destroys the fairness of the original BCNet method. Hence, with the pre-set smallest width $l_s$, all channels indexed by AutoSlim and BCNet with Eq. \eqref{ua_channel} and Eq. \eqref{bc_channel} can be updated as Eq. \eqref{ua_channel_update} and Eq. \eqref{bc_channel_update}, respectively.
\begin{equation}
\I_{UA}(c, l_s) = [1:c],~\forall~c = l_s,...,l. 
\label{ua_channel_update}
\vspace{-4mm}
\end{equation}
\begin{multline}
\I_{BC}(c, l_s) = \I_{UA}^l(c, l_s) \uplus \I_{UA}^r(c, l_s) \\
= [1:c] \uplus [(l-c+1):l], ~\forall~c = l_s,...,l. 
\label{bc_channel_update}
\end{multline}

Since the index range changed with the pre-set smallest width, the cardinality of AutoSlim and BCNet are thus updated from Eq. \eqref{ua_counts} and Eq. \eqref{bc_counts} to Eq. \eqref{ua_counts_update} and Eq. \eqref{bc_counts_update}, respectively, as piecewise functions based on the pre-set $l_s$.
\begin{equation}
    \mbox{Card-UA}(c, l_s) = \left\{
        \begin{array}{l}
            l-c+1, ~\forall~c = l_s,...,l.  \\
            l-l_s+1, ~\forall~c = 1,...,l_s.  \\
        \end{array}
    \right.\label{ua_counts_update}
\end{equation}

\begin{multline}
    \mbox{Card-BC}(c, l_s) = \mbox{Card-UA}(c, l_s) + \mbox{Card-UA}(l+1-c, l_s) \\
    = \left\{
        \begin{array}{l}
            (l-c+1) + (l+1-l-1+c) \\
            = l+1 , ~\forall~c = l_s,...,l.  \\
            (l-l_s+1) + (l+1-l-1+c) \\
            = l+1+c-l_s, ~\forall~c = 1,...,l_s.  \\
        \end{array}
    \right.\label{bc_counts_update}
\end{multline}

As illustrated in Eq. \eqref{ua_counts_update} and Eq. \eqref{bc_counts_update}, since the searched widths are augmented by base widths $l_s$, the searched ones with more channels (\emph{i.e.}, a wider model) contain both of the base and searched channels. In this case, under common one-shot frameworks, for AutoSlim \cite{autoslim} and BCNet in Section \ref{BCNet_supernet}, due to the appearance of the base channels, the cardinality of channels within the searched widths is larger than those in the base widths.

For sufficiently and fairly training the network widths, as depicted in Figure \ref{fig:DB_width}(c), we concatenate two supernets together by overlapping the grouping channels and base channels from the two sides of the supernet by turn in batches in order to statistically balance their assignment in supernet training, namely BCNetV2.

Therefore, BCNetV2 enables the supernets from both the left and right sides to have their own base widths $l_s$. As the maximum searched widths is set as $l$, the overall widths of the BCNetV2 supernet in a layer should be $l+l_s-d$ ($d=(l-l_s)/(K-1)$). Hence, the channels $[1,...,l_s-d]$ and $[l+1,...,l+l_s-d]$ are the base channels of left and right supernets, respectively. As a result, all channels $\I_{BCv2}(c)$ used for evaluating width $c$ in BCNetV2 are indexed with
\begin{multline}
\I_{BCv2}(c) = \I_{UA}^{BCv2-l}(c, l_s) \uplus \I_{UA}^{BCv2-r}(c, l_s) \\
= [1:c] \uplus [(l+l_s-d-c+1):(l+l_s-d)], ~\forall~c = l_s,...,l. 
\label{DB_bc_channel}
\end{multline}

Concretely, the cardinality of each channel within BCNetV2 principle is the sum from both sides of supernets $\N_l^{BCv2}$ and $\N_r^{BCv2}$. In detail, since channels count from the right side within $\N_r^{BCv2}$, the cardinality for the $c$-th channel in left side corresponds to the cardinality of $l+l_s-d-c+1$ channel in the right side with Eq. \eqref{DB_bc_channel}. Therefore, the cardinality of the $c$-th channel in BCv2 principle is
\begin{multline}
    \mbox{Card-BCv2}(c, l_s) \\  
    = \mbox{Card-UA}(c, l_s) + \mbox{Card-UA}(l+l_s-c+1, l_s) \\
    = \left\{
        \begin{array}{l}
            (l-c+1) + (l-(l+l_s-c+1)+1) \\
            = l+1-l_s , ~\forall~c = l_s,...,l.  \\
            (l-l_s+1) + 0 \\
            = l+1-l_s, ~\forall~c = 1,...,l_s.  \\
        \end{array}
    \right.\label{DB_bc_counts}
\end{multline}
when $c$ is smaller than the base channels $l_s$, the cardinality of $c$-th channel of one supernet is not affected by the other side, which amounts to $0$ in Eq. \eqref{DB_bc_counts}. With the updated BCNetV2 in Eq. \eqref{DB_bc_channel}, the cardinality for each channel will always amounts to the same constant value of $l+1-l_s$, and irrelevant with the index of channels with BCNetV2 principle, thus promoting the fairness in term of channel (filter) level and also reduce the redundance of search space with the base widths $l_s$, which enables to fairly rank network widths and boost the search results with BCNetV2.

\subsection{Iteratively updating of BCNetV2 in a memory efficient way}

To train the BCNetV2 supernets, we can adopt the same stochastic way as in Section \ref{stochastic_training}. As formulated in Eq. \eqref{eq7}, for each batch $\B$ and a sampled network width $c$ from the search space $\C$, the training batch $\B$ is supposed to forward simulaneously through both left $\mathcal{N}^*_l(W)$ and right $\mathcal{N}^*_r(W)$ supernets. With this setting, BCNetV2 also has two times memory comparing to AutoSlim \cite{autoslim}, which hampers the usage of our algorithm in edge devices. 

To reduce the memory usage of BCNetV2, we propose to train the supernets in a iterative manner. In detail, we selectively train the left $\mathcal{N}^*_l(W)$ and right $\mathcal{N}^*_l(W)$ supernets with odd and even number of batches, respectively. Therefore, the Eq. \eqref{eq7} is updated as Eq. \eqref{update_eq7}.
\begin{equation}
    \L_{tr}(W,\c;\B)=\left\{
        \begin{array}{l}
            \L_{tr}(\mathcal{N}_l; \c, \B),~\forall~odd~\B  \\
            \L_{tr}(\mathcal{N}_r; \c, \B),~\forall~even~\B  \\
        \end{array}
    \right.\label{update_eq7}
\end{equation}

Moreover, our BCNetV2 can also achieve the fair training of channels and boost the search performance with no more memory usage than other methods \cite{amc,autoslim,tas}. As a result, our BCNetV2 can achieve the same memory usage with more fairly ranking weight sharing paradigm.

\section{A New Benchmark: Channel-Bench-Macro}

For a better comparison between our BCNet and other width search methods, we propose the first open source width benchmark on macro structures with CIFAR-10 dataset, named Channel-Bench-Macro \footnote{https://github.com/xiusu/Channel-Bench-Macro}. The Channel-Bench-Macro consists of two base models of MobileNet and ResNet and both have 16,384 architectures and their test accuracies, parameter numbers, and FLOPs on CIFAR-10.

\begin{figure}[!t]
    \centering
	\begin{subfigure}
        \centering
        \includegraphics[width=1\linewidth]{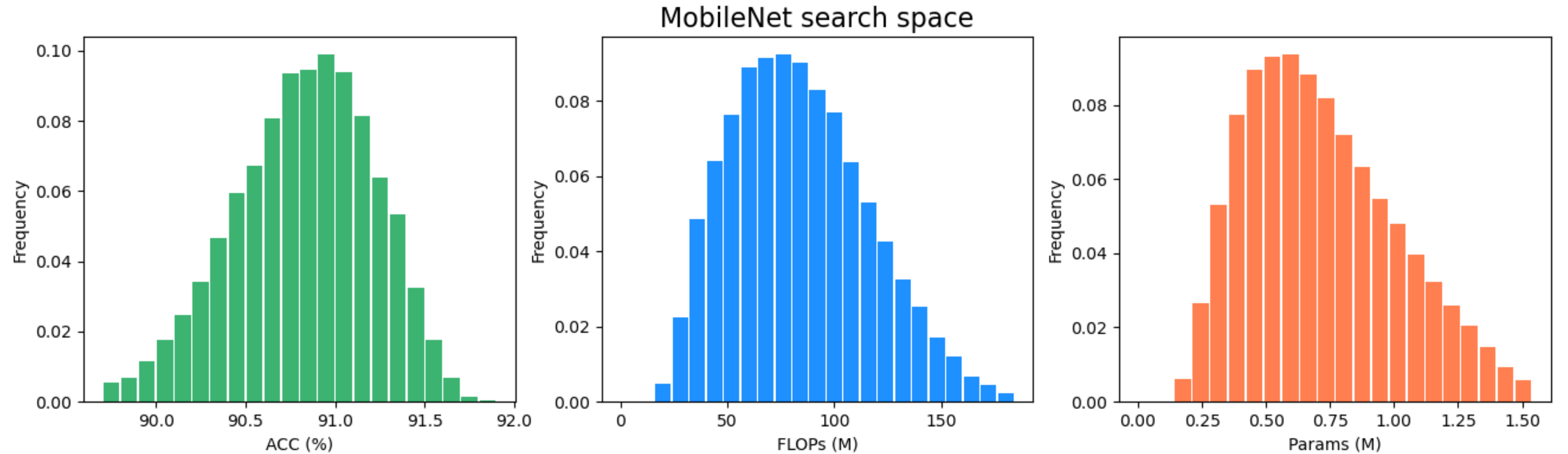}
    \end{subfigure}
	\begin{subfigure}
        \centering
        \includegraphics[width=1\linewidth]{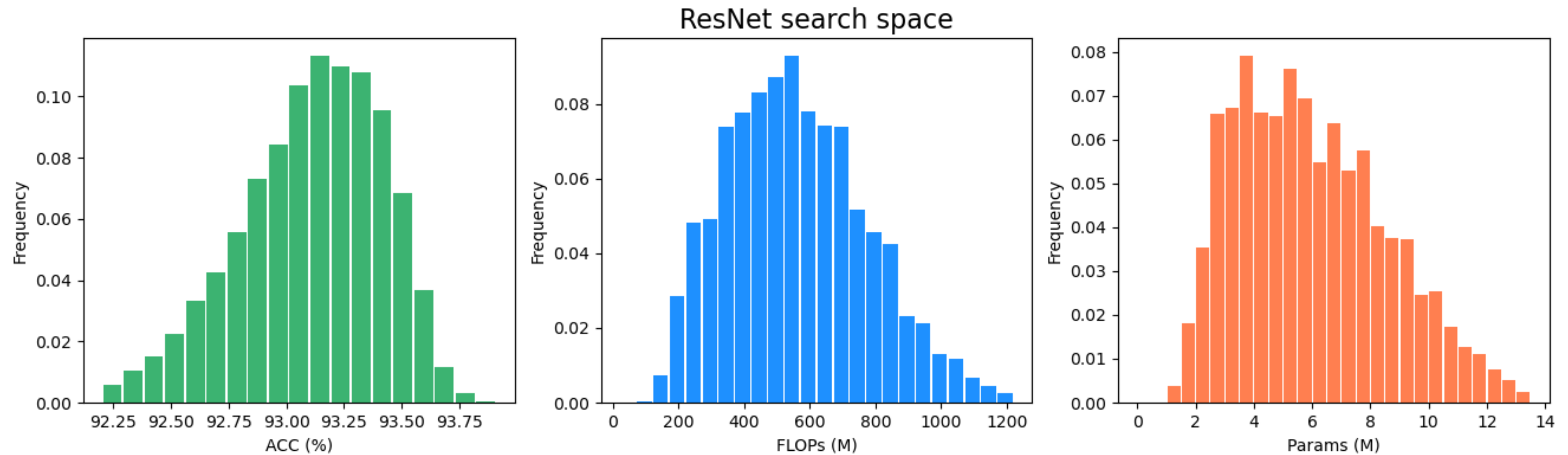}
    \end{subfigure}
    \vspace{-5mm}
    \caption{Histograms of accuracies, FLOPs, and parameters of architectures in Channel-Bench-Macro w.r.t. MobileNet search space and ResNet search space.}\label{fig:widthbench_histograms}
    \vspace{-6mm}
\end{figure}

\subsection{Search space of Channel-Bench-Macro}

For each of the models, the search space of the Channel-Bench-Macro is conducted with 7 searching layers; each layer contains 4 uniformly distributed candidate widths. Thus the overall search space of two models is $4^7\times2=32,768$. For the skipping lines or depthwise related layers, we merge the layers which are required to have the same widths. For ResNet-based architectures in the Channel-Bench-Macro, we follow the same block setting as ResNet34 and ResNet18. We apply the skipping lines for all blocks when the output channels is equal to the input channels.

\begin{table}[h]
    \centering
    \caption{Macro structure of MobileNet search space on Channel-Bench-Macro. ``ratio'' indicates the expansion ration of the MobileNet block, ``channel'' represents the output channel of corresponding block.}
    \vspace{-2mm}
    \begin{tabular}{c|c|c|c|c|c}
        \hline
        n & input & block & channel & ratio & stride \\
        \hline
        1 & $32\times32\times3$ & $3\times3$ conv&  128 & - & 2\\
        2 & $16\times16\times128$ & MB Block & 128 & 6 & 1 \\
        1 & $16\times16\times128$ & MB Block & 256 & 6 & 2 \\
        1 & $8\times8\times256$ & MB Block & 256 & 6 & 1 \\
        1 & $8\times8\times256$ & global avgpool & - & - & - \\
        1 & $1\times1\times256$ & FC & 10 & - & - \\
        \hline
    \end{tabular}
    \vspace{-3mm}
    \label{tab:nasbench_search_space}
\end{table}

\begin{table}[h]
    \centering
    \caption{Macro structure of ResNet search space on Channel-Bench-Macro.``mid\_channel'' indicates the output channel of the first convolution in each block.}
    \vspace{-3mm}
    \begin{tabular}{c|c|c|c|c|c}
        \hline
        n & input & block & channel & mid\_channel & stride \\
        \hline
        1 & $32\times32\times3$ & $3\times3$ conv&  256 & - & 2\\
        2 & $16\times16\times256$ & ResNet Block & 256 & 256 & 1 \\
        1 & $16\times16\times256$ & ResNet Block & 512 & 512 & 2 \\
        2 & $8\times8\times512$ & ResNet Block & 512 & 512 & 1 \\
        1 & $8\times8\times512$ & global avgpool & - & - & - \\
        1 & $1\times1\times512$ & FC & 10 & - & - \\
        \hline
    \end{tabular}
    \vspace{-3mm}
    \label{tab:nasbench_search_space}
\end{table}

\subsection{Benchmarking architectures on Channel-Bench-Macro}

We train all the aforementioned 32,768 architectures isolatedly on the CIFAR-10 dataset. Each architecture is trained with a batch size of 256 and a SGD optimizer, a cosine learning rate strategy that decays 60 epochs is adopted with an initial value 0.1. And the weight decay is set to $5\times10^{-4}$. We train each architectures three times with different random seeds and report their mean accuracies on the test set.

\subsection{Statistics on Channel-Bench-Macro training results}

We analyze the distribution of parameters, FLOPs, and accuracies of all architectures in the Channel-Bench-Macro, which are depicted in Figure \ref{fig:widthbench_histograms}. From the results, we can see that, with channels uniformly sampled, the FLOPs, parameters, and accuracies of models perform nearly to a skewed Gaussian distribution. The accuracies benefit less from the increase of FLOPs or parameters, this makes the width search algorithms hard to rank large architectures accurately.

\subsection{Rank correlation of parameters and FLOPs with accuracies on Channel-Bench-Macro}

We measure the rank correlation between parameters and FLOPs~\wrt~accuracies on Channel-Bench-Marco. The results summarized in Table~\ref{tab:widthbench_correlation_flops_params} show that the FLOPs has higher rank correlation coefficients than parameters. It indicates that the increment of FLOPs contributes more to accuracy than increasing parameters. 

\begin{table}[h]
	\centering
	\vspace{-3mm}
	\caption{Rank correlations of parameters and FLOPs with accuracies on Channel-Bench-Macro with MobileNet like architectures. }  
	\vspace{-3mm}
	\label{tab:widthbench_correlation_flops_params}
	\begin{tabular}{c||c|c|c}
		\hline
		type & Pearson (\%) & Spearman rho (\%) & Kendall tau (\%) \\
		\hline	
		Params & 60.3 & 60.3 & 42.6 \\ 
		FLOPs & 72.0 & 72.1 & 52.6 \\
		\hline
	\end{tabular}
	\vspace{-5mm}
\end{table}

\begin{table}[h]
	\centering
	\caption{Rank correlations of parameters and FLOPs with accuracies on Channel-Bench-Macro with ResNet like architectures. }  
	\label{tab:widthbench_correlation_flops_params}
	\vspace{-3mm}
	\begin{tabular}{c||c|c|c}
		\hline
		type & Pearson (\%) & Spearman rho (\%) & Kendall tau (\%) \\	
		\hline	
		Params & 61.9 & 64.1 & 45.9 \\ 
		FLOPs & 75.6 & 78.7 & 59.4 \\
		\hline
	\end{tabular}
	\vspace{-3mm}
\end{table}

\subsection{Meaning of Channel-Bench-Macro.} 

The key challenge of the width search algorithm lies in the evaluation and ranking reliability of the supernet, which can be reflected by the ranking correlation between the evaluation performance of all the architectures in the supernet and their actual performance. However, currently, there is no public benchmark for width (channel) search space that provides performance of all networks by retraining from scratch to our best knowledge. Therefore, we construct the Channel-Bench-Macro with the controlled size of the search space for the experiments of verifying our method's effectiveness in terms of searching efficiency and performance and also providing a way of fair comparison of different algorithms.

\section{Experimental Results}

In this section, we conduct extensive experiments on the ImageNet and CIFAR-10 datasets to validate the effectiveness of our algorithm. For all structures, we search on the reduced space $\C_K$ with default $K=20$.  Note that most pruning methods do not report their results by incorporating the knowledge distillation (KD) \cite{distilling,du2020agree} improvement in retraining except for MobileNetV2. Thus in our method, except for MobileNetV2, we also do not include KD in final retraining.

\textbf{Comparison methods.} We include multiple competing pruning, network width search methods and NAS models for comparison, such as DMCP \cite{dmcp}, TAS \cite{tas}, AutoSlim \cite{autoslim}, MetaPruning \cite{metapruning}, AMC \cite{amc}, DCP \cite{dcp}, LEGR \cite{legr}, CP \cite{cp}, AutoPruner \cite{autopruner}, SSS \cite{sss}, EfficientNet-B0 \cite{efficientnet} and ProxylessNAS \cite{proxylessnas}. Moreover, we also consider two vanilla baselines. \textit{Uniform}: we shrink the width of each layer with a fixed factor to meet FLOPs budget. \textit{Random}: we randomly sample 20 networks under FLOPs constraint, and train them by 50 epochs, then we continue training the one with the highest performance and report its final result.

\subsection{Implementation details of training recipe} 

In this section, we present the training details of our BCNet \wrt~experiments on various models. In detail, we search on the reduced space $\C_K$ with default $K=20$. During training, except for EfficientNet-B0 and ProxylessNAS, we use SGD optimizer with momentum 0.9 and nesterov acceleration. As for EfficientNet-B0 and ProxylessNAS, we adopt RMSprop optimizer for searching optimal network width. 

\subsubsection{Training recipe of ResNet50, ResNet34, ResNet18, MobileNetV2, and VGGNet.}
For ResNet models, we follow the same training recipe as TAS \cite{tas}. In detail, we use a weight decay of $10^{-4}$ and batch size of 256; and we train the model by 120 epochs with the learning rate annealed with cosine strategy from initial value 0.1 to $10^{-5}$. For MobileNetV2 and VGGNet, we set weight decay to $5\times10^{-5}$ and $10^{-4}$, respectively. Besides, for MobileNetV2, we adopt the batch size of 256, and the learning rate is annealed with a cosine strategy from initial value $0.1$ to $10^{-5}$. For VGGNet, we train it for 400 epochs using a batch size of 128; the learning rate is initialized to 0.1 and divided by 10 at 160-th, 240-th epoch. Moreover, we note that most pruning methods do not report their results by incorporating the knowledge distillation (KD) \cite{distilling} improvement in retraining except for MobileNetV2. Thus in our method, except for MobileNetV2, we do not include KD in final retraining for a more fair comparison of performance. All experiments are implemented with PyTorch on NVIDIA V100 GPUs.

\subsubsection{Training recipe of EfficientNet-B0 and ProxylessNAS.} 
We use the same training strategies for both EfficientNet-B0 and ProxylessNAS. In detail, we train both models for 300 epochs using a batch size of 1024; the learning rate is initialized to 0.128 and decayed by 0.963 for every 3 epochs. Besides, the first 5 training epochs are adopted as warm-up epochs, and the weight decay is set to 1$\times 10^{-5}$.

\begin{table}[!t]
	\centering
	\footnotesize
	\caption{Searching results of EfficientNet-B0 and ProxylessNAS on ImageNet dataset.}
	\label{Experiments_Efficientnetb0}
	\vspace{-2mm}
	{\begin{tabular}{c|c|c|c|c}
			\hline
			\multicolumn{5}{c}{EfficientNet-B0} \\ \hline
			Groups&Methods&Param&Top-1&Top-5  \\ \hline
			\multirow{3}*{385M}&Uniform & 5.3M & 76.88\% & 92.64\% \\
			&Random & 5.1M & 76.37\% & 92.25\% \\
			& BCNet & 6.9M & 77.36\% & 93.17\% \\ 
			& BCNetV2 & 6.6M & \textbf{77.53\%} & 93.31\% \\ \hline
			\multirow{3}*{192M}&Uniform & 2.7M & 74.26\% & 92.24\% \\
			&Random & 2.9M & 73.82\% & 91.86\% \\
			& BCNet & 3.8M & 74.92\% & 92.06\% \\ 
			& BCNetV2 & 3.8M & \textbf{75.24\%} & 92.28\% \\ \hline
			\multicolumn{5}{c}{ProxylessNAS} \\ \hline
			Groups&Methods&Param&Top-1&Top-5  \\ \hline
			\multirow{3}*{320M} & Uniform & 4.1M & 74.62\% & 91.78\% \\
			& Random & 4.3M & 74.16\% & 91.23\% \\
			&BCNet & 5.4M & 75.07\% & 91.97\% \\ 
			&BCNetV2 & 5.6M & \textbf{75.26\%} & 92.13\% \\ \hline
			\multirow{3}*{160M} & Uniform & 2.2M & 71.16\% & 89.49\% \\
			&Random & 2.5M & 70.89\% & 89.12\% \\
			&BCNet & 2.9M & 71.87\% & 89.96\% \\
			&BCNetV2 & 2.8M & \textbf{72.14\%} & 90.14\% \\ \hline
	\end{tabular}}	
	\vspace{-4mm}
\end{table}

\subsection{Results on ImageNet dataset} 
ImageNet dataset contains 1.28M training images and 50K validation images from 1K classes. In specific, we report the accuracy on the validation dataset as \cite{slimming,autoslim}, and the original model takes as the supernet while for the 1.0$\times$ FLOPs of all models, the supernet refers to a 1.5$\times$ FLOPs of original model by uniform width scaling. To verify the performance on both heavy and light models, we search on the ResNet models and MobileNetV2 with different FLOPs budgets. In our experiment, the original ResNet50 and MobileNetV2 has 25.5M, 3.5M parameters and 4.1G, 300M FLOPs with 77.5\%, 72.6\% Top-1 accuracy, respectively. 

As shown in Table \ref{Experiments_Imagenet}, our BCNet achieves the highest accuracy on ResNet models and MobileNetV2 \wrt~different FLOPs, which indicates the superiority of our BCNet to other pruning methods. For example, our 3G FLOPs ResNet50 decreases only 0.2\% Top-1 accuracy compared to the original model, which exceeds AutoSlim \cite{autoslim} and MetaPruning \cite{metapruning} by 1.3\% and 1.1\%. While for MobileNetV2, our 207M MobileNetV2 exceeds the state-of-the-art AutoSlim, MetaPruning by 0.4\%, 1.1\%, respectively. In addition, our BCNet even surpasses other algorithms more on tiny MobileNetV2 (105M) with 68\% Top-1 accuracy and exceeds MetaPruning by 3.0\%.

To further demonstrate the effectiveness of our BCNet on highly efficient models, we conduct searching on the NAS-based models EfficientNet-B0 and ProxylessNAS. The original EfficientNet-B0 (ProxylessNAS) has 5.3M (4.1M) parameters and 385M (320M) FLOPs with 76.88\% (74.62\%) Top-1 accuracy, respectively. As shown in Table \ref{Experiments_Efficientnetb0}, although the increase of performance is not as signicant as in Table \ref{Experiments_Imagenet}, our method can still boost the NAS-based models by more than 0.4\% on Top-1 accuracy.

\begin{table}[t]
	\centering
	\caption{Performance comparison of MobileNetV2 and VGGNet on CIFAR-10.}
	\label{Experiment_Cifar10}
	\vspace{-2mm}
	\scriptsize
	{\begin{tabular}{c|c|ccc}
			\hline
			\multicolumn{5}{c}{MobileNetV2} \\ \hline
			Groups&Methods&FLOPs&Params&accuracy  \\ \hline
			\multirow{4}*{200M} & DCP \cite{dcp} & 218M & - & 94.69\% \\
			& Uniform & 200M & 1.5M & 94.57\% \\
			& Random & 200M & - & 94.20\% \\
			& \textbf{BCNet} & 200M & 1.5M & \textbf{95.44\%}\\ \cline{1-5}
			
			\multirow{4}*{146M} & MuffNet \cite{muffnet} & 175M & - & 94.71\% \\ 
			& Uniform & 146M & 1.1M & 94.32\% \\
			& Random & 146M & - & 93.85\% \\
			& \textbf{BCNet} & 146M & 1.2M & \textbf{95.42\%} \\ \cline{1-5}
			
			\multirow{5}*{44M} & AutoSlim \cite{autoslim} & 88M & 1.5M & 93.20\% \\
			& AutoSlim \cite{autoslim} & 59M & 0.7M & 93.00\% \\
			& MuffNet \cite{muffnet} & 45M & - & 93.12\% \\
			& Uniform & 44M & 0.3M & 92.88\% \\
			& Random & 44M & - & 92.31\% \\
			& \textbf{BCNet} & 44M  & 0.4M & \textbf{94.42\%} \\ \cline{1-5}
			
			\multirow{4}*{28M} & AutoSlim \cite{autoslim} & 28M & 0.3M & 92.00\%\\
			& Uniform & 28M & 0.2M & 92.37\% \\
			& Random & 28M & - & 91.69\% \\
			& \textbf{BCNet} & 28M & 0.2M & \textbf{94.02\%} \\ \hline
			\multicolumn{5}{c}{VGGNet} \\ \hline
			Groups&Methods&FLOPs&Params&accuracy  \\ \hline
			\multirow{5}*{200M} & Sliming \cite{slimming} & 199M & 10.4M & 93.80\% \\
			& DCP \cite{dcp} & 199M & 10.4M & 94.16\% \\
			& Uniform & 199M & 10.0M & 93.45\%  \\
			& Random & 199M & - & 93.02\% \\ \cline{1-5}
			& \textbf{BCNet} & 197M & 3.1M & \textbf{94.36\%}  \\ 
			\multirow{8}*{100M$+$} & Uniform & 185M & 9.3M & 93.30\% \\
			& Random & 185M & - & 92.71\%  \\
			& \textbf{BCNet} & 185M & 6.7M & \textbf{94.14\%} \\ 
			& CP \cite{cp} & 156M & 7.7M & 93.67\%  \\
			& Multi-loss \cite{multi} & 140M & 5.5M & 94.05\% \\
			& Uniform & 138M & 6.8M & 93.14\% \\
			& Random & 138M & - & 92.17\% \\
			& \textbf{BCNet} &  138M & 3.3M & \textbf{94.09\%} \\ \cline{1-5}
			\multirow{5}*{77M} & CGNets \cite{cgnet} & 91.8M & - & 92.88\% \\
			& Uniform & 77.0M & 3.9M & 92.38\% \\
			& Random & 77.0M & - & 91.72\%  \\
			& \textbf{BCNet} & 77.0M & 1.2M & \textbf{93.53\%} \\
			& CGNet \cite{cgnet} & 61.4M & - & 92.41\% \\ \hline
	\end{tabular}}	
	\vspace{-3mm}
\end{table}

\begin{table*}[!t]
	\centering
	\scriptsize
	\caption{Performance report of BCNetV2~\wrt~baseline methods with ResNet50 and MobileNetV2 on ImageNet. Methods with "*" denotes  that the results are reported with knowledge distillation. BCNet$^\dagger$ indicates that the search is conducted with the shrinked search space with base width $l_s$.}
	\label{DB_Experiments_Imagenet}
	\vspace{-2mm}
	{\begin{tabular}{c|c|cc|cc||c|c|cc|cc}
			\hline
			\multicolumn{6}{c||}{ResNet50} & \multicolumn{6}{c}{MobileNetV2}\\ \hline
			FLOPs level&Methods&FLOPs&Parameters&Top-1&Top-5&FLOPs level&Methods&FLOPs&Parameters&Top-1&Top-5  \\ \hline
			\multirow{14}*{2G} & GBN \cite{gbn} & 2.4G & 31.8M & 76.2\% & 92.8\% &  \multirow{8}*{305M(1.5$\times$)} & AutoSlim* \cite{autoslim} & 305M & 5.7M & 74.2\% & - \\ 
			& SFP \cite{sfp} & 2.4G & - & 74.6\% & 92.1\% && Uniform & 305M & 3.6M & 72.7\% & 90.7\% \\
			& LEGR \cite{legr} & 2.4G & - & 75.7\% & 92.7\% && Random & 305M & - & 71.8\% & 90.2\% \\
			& FPGM \cite{fpgm} & 2.4G & - & 75.6\% & 92.6\% && BCNet & 305M & 4.8M & 73.9\% & 91.5\% \\ 
			& TAS* \cite{tas} & 2.3G & - & 76.2\% & 93.1\% && BCNet* & 305M & 4.8M & 74.7\% & 92.2\% \\ 
			& DMCP \cite{dmcp} & 2.2G & - & 76.2\% & - && BCNet$^\dagger$ & 305M & 4.7M & 73.6\% & 91.8\% \\ 
			& MetaPruning \cite{metapruning} & 2.0G & - & 75.4\% & - && BCNetV2 & 305M & 4.9M & 74.2\% & 91.7\% \\ 
			& AutoSlim* \cite{autoslim} & 2.0G & 20.6M & 75.6\% & - && BCNetV2* & 305M & 4.9M & 75.0\% & 92.5\% \\ \cline{7-12}
			& Uniform & 2.0G & 13.3M & 75.1\% & 92.7\% & \multirow{13}*{150M} & TAS* \cite{tas} & 150M & - & 70.9\% & - \\
			& Random & 2.0G & - & 74.6\% & 92.2\% && LEGR \cite{legr} & 150M & - & 69.4\% & - \\
			& BCNet & 2.0G & 18.4M & 76.9\% & 93.3\% && AMC* \cite{amc} & 150M & - & 70.8\% & - \\
			& BCNet$^\dagger$ & 2.0G & 18.3M & 76.7\% & 93.1\% && Uniform & 150M & 2.0M & 69.3\% & 88.9\% \\
			& \textbf{BCNetV2} & 2.0G & 17.8M & \textbf{77.3\%} & \textbf{93.4\%} && Random & 150M & - & 68.8\% & 88.7\% \\ \cline{1-6}
			\multirow{8}*{1G} & AutoPruner \cite{autopruner} & 1.4G & - & 73.1\% & 91.3\% && BCNet & 150M & 2.9M & 70.2\% & 89.2\% \\
			& MetaPruning \cite{metapruning} & 1.0G & - & 73.4\% & -  && BCNet* & 150M & 2.9M & 71.2\% & 89.6\% \\
			& AutoSlim \cite{autoslim} & 1.0G & 13.3M & 74.0\% & - && BCNet$^\dagger$ & 150M & 2.8M & 69.7\% & 88.9\% \\
			& Uniform & 1.0G & 6.9M & 73.1\% & 91.8\% && \textbf{BCNetV2} & 150M & 2.6M & \textbf{70.7\%} & \textbf{89.3\%} \\
			& Random & 1.0G & - & 72.2\% & 91.4\% && \textbf{BCNetV2*} & 150M & 2.6M & \textbf{71.9\%} & \textbf{90.0\%}  \\ \cline{7-12}
			& BCNet & 1.0G & 12M & 75.2\% & 92.6\% & \multirow{9}*{50M} & MuffNet \cite{muffnet} & 50M & - & 50.3\% & - \\ 
			& BCNet$^\dagger$ & 1.0G & 11.7M & 74.9\% & 92.5\% && MetaPruning \cite{metapruning} & 43M & - & 58.3\% & - \\
			& \textbf{BCNetV2} & 1.0G & 11.6M & \textbf{75.7\%} & \textbf{93.0\%} && Uniform & 50M & 0.9M & 59.7\% & 82.0\% \\ \cline{1-6}
			\multirow{6}*{570M} & AutoSlim* \cite{autoslim} & 570M & 7.4M & 72.2\% & - && Random & 50M & - & 57.4\% & 81.2\% \\
			& Uniform & 570M & 6.9M & 71.6\% & 90.6\%  && BCNet & 50M & 1.6M & 62.7\% & 83.7\% \\
			& Random & 570M & - & 69.4\% & 90.3\% && BCNet* & 50M & 1.6M & 63.8\% & 84.6\% \\
			& BCNet & 570M & 12.0M & 73.2\% & 91.1\% && BCNet$^\dagger$ & 50M & 1.4M & 62.4\% & 83.5\% \\
			& BCNet$^\dagger$ & 570M & 11.8M & 73.0\% & 90.8\% && \textbf{BCNetV2} & 50M & 1.5M & \textbf{63.3\%} & \textbf{83.9\%} \\
			& BCNetV2 & 570M & 12.1M & 73.8\% & 91.5\%   && \textbf{BCNetV2*} & 50M & 1.5M & \textbf{64.4\%} & \textbf{84.7\%} \\ \hline
	\end{tabular}}	
	\vspace{-3mm}
\end{table*}

\textbf{Further boosting the searched results with BCNetV2.} To evaluate the performance of the proposed BCNetV2, with the base widths $l_s$ adopted as 5 groups (K=20 groups in total), we conduct the search with the MobileNetV2 and ResNet50 backbones on ImageNet dataset. As reported in Table \ref{DB_Experiments_Imagenet}, the searched 150M MobileNetV2 and 2G ResNet50 achieves 71.9\% and 77.3\% on Top-1 accuracy, which is 0.7\% and 0.4\% surpasses than the original BCNet, respectively. Compare to the other methods, our BCNetV2 achieves 1.0\%$\sim$5.1\% than other baseline methods, which shows the superiority of our BCNetV2. More ablations about BCNetV2 are discussed in section \ref{ab_dbBCNet}.

\begin{table}
	\centering
	\caption{Rank correlations between each supernet and NAS-Bench-Macro. }  
	\label{tab:channelbench_correlation}
	\vspace{-2mm}
	\begin{tabular}{c||c|c|c}
		\hline
		\multicolumn{4}{c}{MobileNet} \\ \hline
		Method & Pearson (\%)& Spearman (\%) & Kendall tau(\%) \\	
		\hline	
		TAS & 76.2 & 75.8 & 58.7 \\ 
		AutoSlim & 79.5 & 79.3 & 60.4 \\
		BCNet & 81.7 & 81.9 & 65.2\\ \hline\hline
        \multicolumn{4}{c}{ResNet} \\ \hline
		Method & Pearson (\%)& Spearman (\%) & Kendall tau(\%) \\	
		\hline	
		TAS & 78.7 & 78.1 & 60.6  \\ 
		AutoSlim & 80.3 & 80.1 & 61.2 \\
		BCNet & 84.9 & 85.7 & 68.4\\ \hline
	\end{tabular}
	\vspace{-4mm}
\end{table}

\subsection{Results on CIFAR-10 dataset}
We also examine the performance of MobileNetV2 and VGGNet on the moderate CIFAR-10 dataset, which has 50K training and 10K testing images with size 32$\times$32 of 10 categories. Our original VGGNet (MobileNetV2) has 20M (2.2M) parameters and 399M (297M) FLOPs with accuracy of 93.99\% (94.81\%).

As shown in Table \ref{Experiment_Cifar10}, our BCNet still enjoys great advantages in various FLOPs levels. For instance, our 200M MobileNetV2 can achieve 95.44\% accuracy, which even outperforms the original model by 0.63\%. Moreover, even with super tiny size (28M), our BCNet can still have 94.02\% accuracy, which surpasses the state-of-the-art AutoSlim\cite{autoslim} by 2.0\%. As for VGGNet, our BCNet is capable of outperforming those competing channel pruning methods DCP \cite{dcp} and Slimming \cite{slimming} by 0.20\% and  0.56\% with 2$\times$ acceleration rate.

\subsection{Rank correlations between each search method on Channel-Bench-Marco}
\begin{figure}[t]
	\centering
	\includegraphics[width=0.6\linewidth]{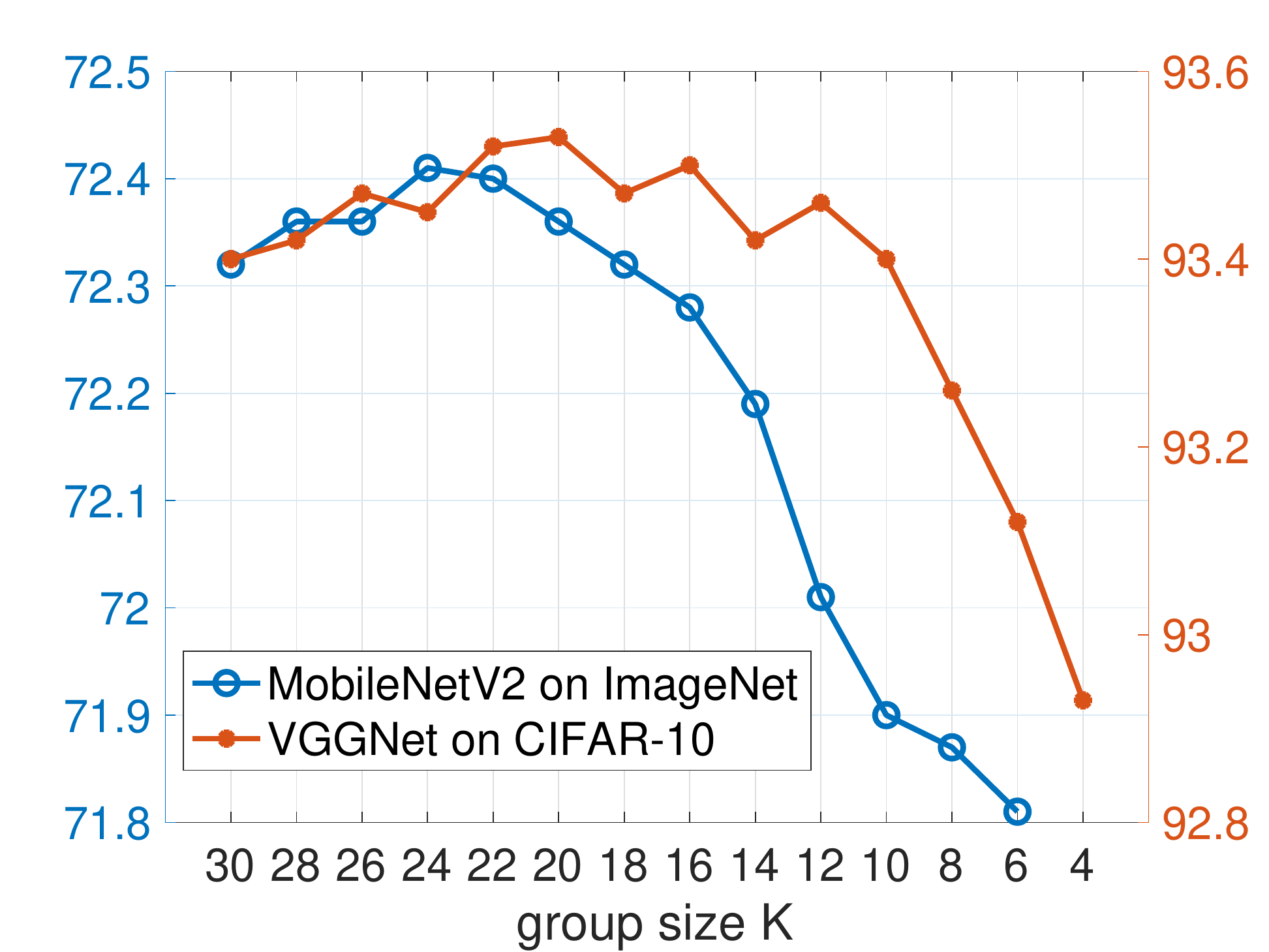}
	\vspace{-2mm}
	\caption{Accuracy performance of the searched network with different group size $K$ of the search space.}
	\label{ratio_list}
	\vspace{-5mm}
\end{figure}
\textbf{Supernet training.} For each width search method, with a batch size of 256, the supernet is trained using a SGD optimizer with 0.9 momentum. A cosine annealing strategy is adopted with an initial learning rate 0.1, which decays 300 epochs. For fully training each subnet, we do not apply weight decay during the supernet training.

In BCNet, during supernet training, the channels are fairly trained and each architecture in search space is got the same probability be trained. Thus, the supernet trained with BCNet weight sharing paradigm should have a more accurate ranking on architectures~\wrt~search space. We adopt experiments to investigate the ranking confidences of BCNet supernet and other baseline methods \cite{tas,autoslim} with the same supernet training strategies on Channel-Bench-Marco. Concretely, we measure the ranking correlation coefficients between the validation accuracies on weight sharing subnets and corresponding ground-truth performances in Channel-Bench-Macro. With the coefficients of Kendall tau, Spearman, and Pearson, we present the results in Table \ref{tab:channelbench_correlation}. From Table \ref{tab:channelbench_correlation}, our BCNet shows more promising results in coefficients~\wrt~MobileNet and ResNet backbones on Channel-Bench-Marco.

\subsection{Ablation studies}
\label{ablation}

\subsubsection{Effect of BCNetV2 as a supernet.} To validate the effectiveness of our proposed supernet BCNetV2, we search the ResNet50, MobileNetV2, EfficientNet-B0 and ProxylessNAS on ImageNet dataset with 2$\times$ acceleration. Our default baseline supernet is that adopted by AutoSlim \cite{autoslim}, which follows unilateral augmented principle to evaluate a network width. As the results in Table \ref{BCNet_analysis} shows, under the greedy search, only using our BCNet evaluation mechanism (second line) can enjoy a gain of 0.27\% to 0.66\% Top-1 accuracy. When searching with evolutionary algorithms, the gain still reaches at 0.28\% to 0.35\% Top-1 accuracy on various models. These exactly indicates using BCNet as supernet could boost the evaluation and searching performance. As for the complementary training strategy, we can see that it enables to boost our BCNet by improving the MobileNetV2 (ResNet50) from 69.92\% (76.41\%) to 70.04\% (76.56\%) on Top-1 accuracy. Note that greedy search without BCNet supernet amounts to AutoSlim, we can further indicate the superiority of our method to AutoSlim with achieved Top-1 accuracy 70.20\% (76.90\%) vs 69.52\% (75.94\%) on MobileNetV2 (ResNet50). Moreover, with the proposed BCNetV2, \ie, ``base widths'' in Table \ref{BCNet_analysis}, the performance of searched architectures can be further boosted by 0.27$\sim$0.51\% on Top-1 accuracy.



\begin{figure}[t]
	\centering
	\includegraphics[width=1\linewidth]{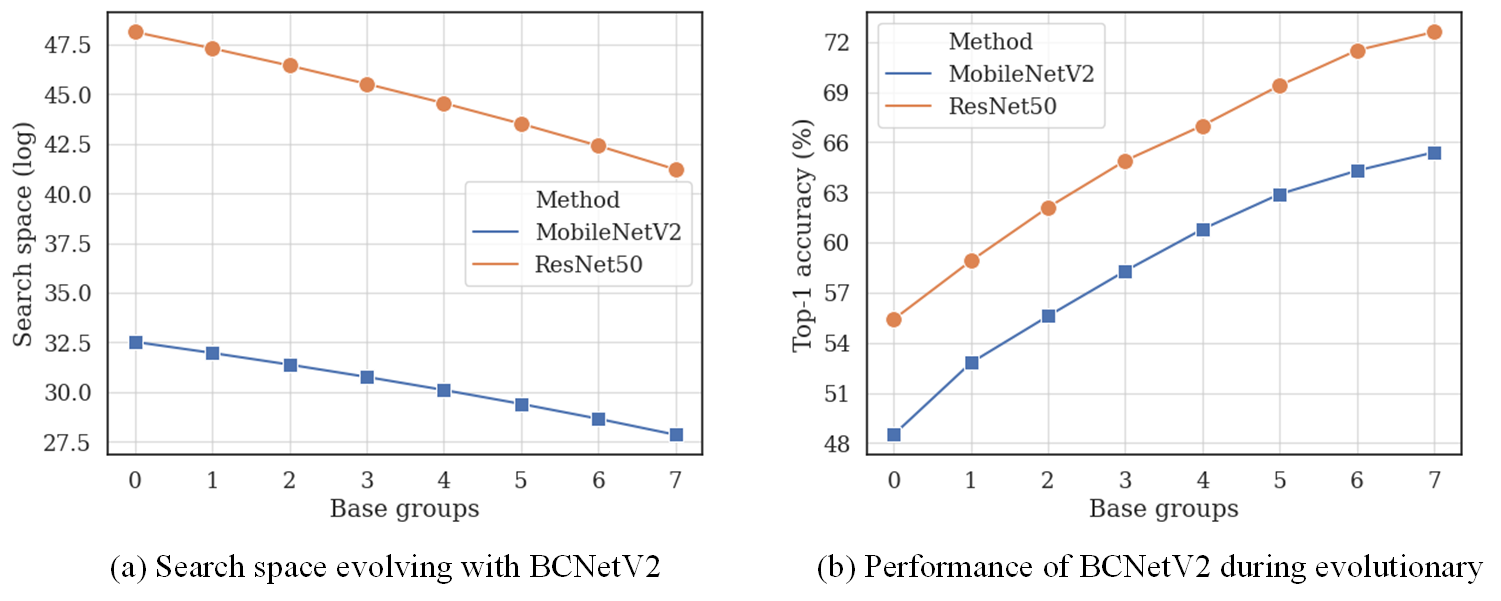}
	\vspace{-2mm}
	\caption{(a) Search space evolving with BCNetV2~\wrt~base widths $l_s$ with MobileNetV2 and ResNet50 on ImageNet dataset. The y-axis takes the logarithm to base 10 to show the order of magnitude change in the data. (b)Top-1 accuracy of best performance widths with BCNetV2 during evolutionary~\wrt~different base widths $l_s$ with MobileNetV2 and ResNet50 on ImageNet dataset.}
	\label{supernet_performance}
	\vspace{-5mm}
\end{figure}

\begin{figure}[!t]
    \centering
	\begin{subfigure}
        \centering
        \includegraphics[width=0.48\linewidth]{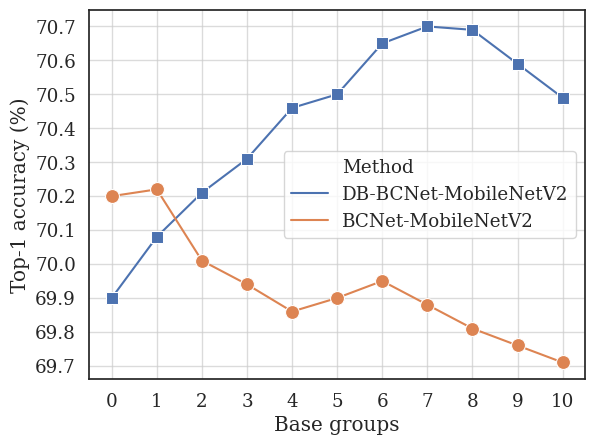}
    \end{subfigure}
	\begin{subfigure}
        \centering
        \includegraphics[width=0.48\linewidth]{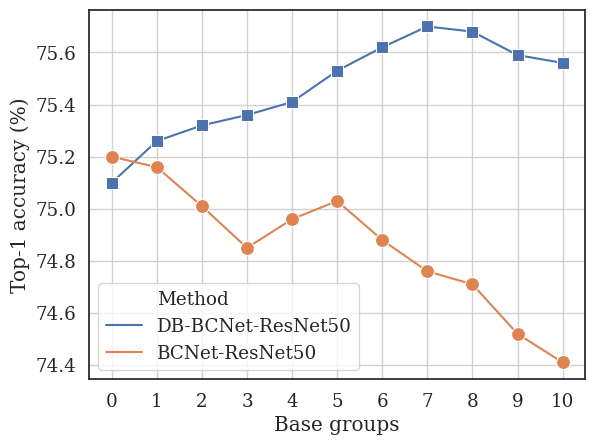}
    \end{subfigure}
    \vspace{-3mm}
    \caption{Training from scratch performance of searched network width of BCNet and BCNetV2~\wrt~different base widths with MobileNetV2 and ResNet50 search space. The FLOPs budget of MobileNetV2 and ResNet50 is set to 150M and 1.0G, respectively.}\label{DB_performance}
    \vspace{-4mm}
\end{figure}

\begin{table*}[t]
	\caption{Performance of searched MobileNetV2 (150M FLOPs), ResNet50 (2G FLOPs), EfficientNet-B0 (192M FLOPs) and ProxylessNAS (160M FLOPs) on ImageNet dataset with different supernet and searching methods. Note that the last line method indicates the recipe of BCNetV2.}
	\label{BCNet_analysis}
	\centering
	\scriptsize
	\begin{tabular}{|c|c|c|c|c|c||cc|cc|cc|cc|} \hline
		\multicolumn{3}{|c|}{evaluator} & \multicolumn{3}{c||}{searching}&  \multicolumn{8}{c|}{models} \\ \cline{1-13} 
		BCNet & complementary & base & greedy & \multicolumn{2}{c||}{evolutionary}& \multicolumn{2}{c|}{MobileNetV2} & \multicolumn{2}{c|}{ResNet50}&\multicolumn{2}{c|}{EfficientNet-B0}& \multicolumn{2}{c|}{ProxylessNAS}\\ \cline{4-13} 
		supernet &training&widths&search&random & prior & Top-1 & Top-5&Top-1 & Top-5&Top-1 & Top-5&Top-1 & Top-5\\ \hline 
		& & & \checkmark& & & 69.52\% & 88.91\% & 75.64\% & 92.90\% & 74.02\% & 91.58\% & 70.97\% & 89.43\% \\
		\checkmark & & &\checkmark& & & 69.87\% & 88.99\% & 76.30\% & 93.16\% & 74.39\% & 91.66\% & 71.24\% & 89.57\%  \\
		\checkmark & \checkmark & &\checkmark& & & \textbf{69.91\%} & 89.02\% & \textbf{76.42\%} & 93.19\% & \textbf{74.51\%} & 91.78\% & \textbf{71.33\%} & 89.62\%\\ \hline
		& & & & \checkmark &  & 69.64\% & 88.85\% & 76.12\% & 92.95\% & 74.35\% & 91.54\% & 71.13\% &  89.49\% \\
		\checkmark & & & & \checkmark &  & 69.92\% & 88.91\% & 76.41\% & 93.12\% & 74.63\% & 91.93\% & 71.48\% & 89.69\%  \\
		\checkmark & \checkmark & & & \checkmark &  & 70.04\% & 89.02\% & 76.56\% & 93.21\% & 74.73\% & 91.85\% & 71.62\% & 89.73\% \\
		\checkmark & \checkmark & & &  & \checkmark & 70.20\% & 89.10\% & 76.90\% & 93.30\% & 74.92\% & 92.06\% & 71.87\% & 89.96\% \\ 
		\checkmark & \checkmark & \checkmark & &  & \checkmark & \textbf{70.71\%} & 89.34\% & \textbf{77.28\%} & 93.37\% & \textbf{75.24\%} & 92.28\% & \textbf{72.14\%} & 90.14\% \\ \hline
	\end{tabular} 
	\vspace{-4mm}
\end{table*}

\begin{figure*}[t]
	\centering
	\includegraphics[angle=90,width=1.00\linewidth]{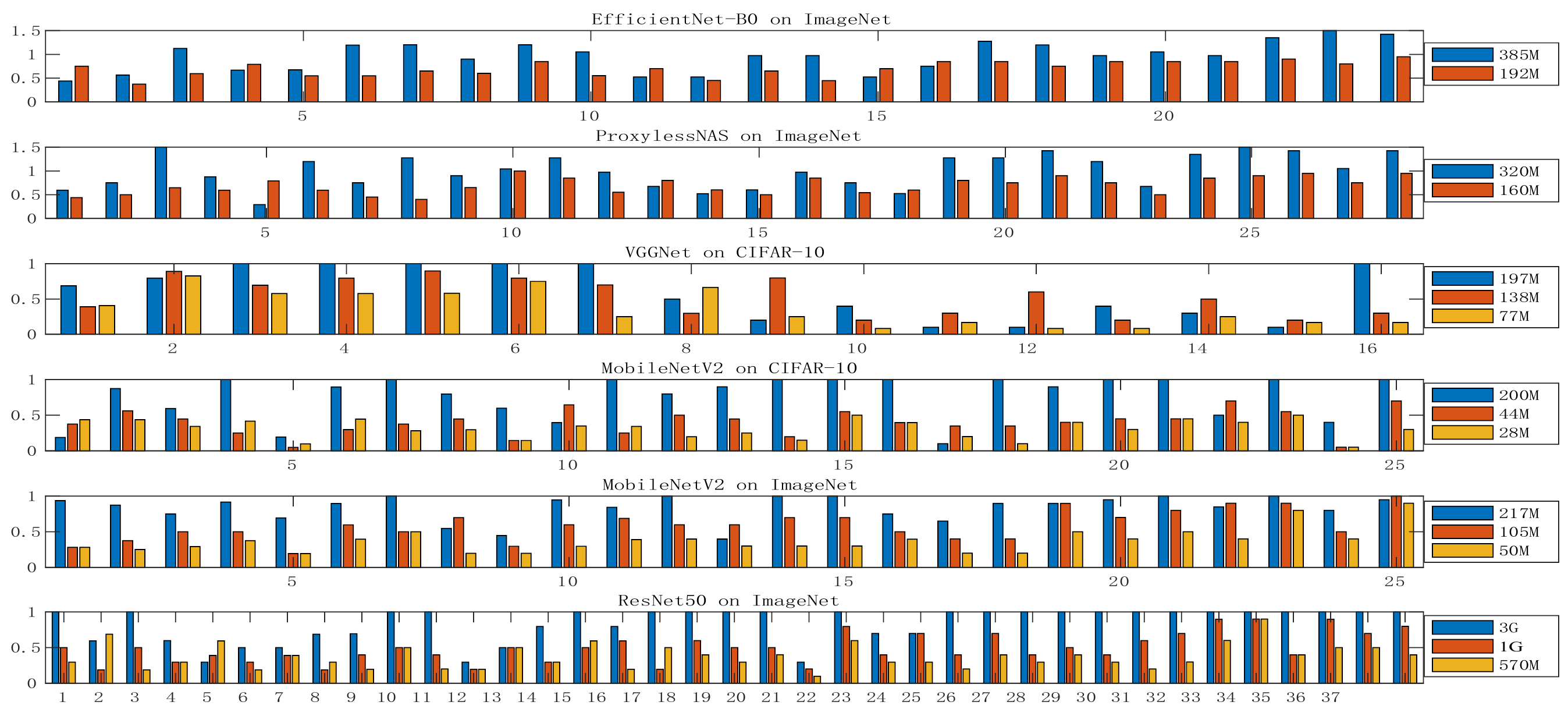}
	\vspace{-6mm}
	\caption{Visualization of searched networks \wrt~different FLOPs. The vertical axis means the ratio of retained channel number compared to that of original networks at each layer.}
	\label{visualization}
	\vspace{-5mm}
\end{figure*}

\subsubsection{Effect of search space.}
We adopt the grouped search space $\C_K$ to reduce its complexity. To investigate the effect of search space, we searched VGGNet on CIFAR-10 dataset and MobileNetV2 on ImageNet dataset with various group size $K$. As in Figure \ref{ratio_list}, our method achieves the best performance in most cases around our default value $K=20$. In addition, we noticed that when the group size $K$ is small, the performance of searched network will increase with $K$ growing larger. This is because group size $K$ determines the size of search space $\C_K$, and larger $K$ induces larger search space. In this way, the obtained network width will be closer to the Oracle optimal width, with higher accuracy achieved accordingly. In addition, the performance tends to be stable when the group size lies in $[14:22]$ but decreases afterwards, which implies the searching space might be  too large for searching an optimal width.

\subsubsection{Ablations of BCNetV2~\wrt~base widths $l_s$} \label{ab_dbBCNet}

\textbf{Search space evolving~\wrt~base widths $l_s$.} With a pre-set base widths $l_s$, the search space of BCNet $K^l$ can be shrinking to $(K-l_s)^l$, where $K$ denotes the grouping number and $l$ indicates the maximum search layers. Take MobileNetV2 as a example, if $l_s=7$, the search space can be reduced from $3.4\times10^{32}$ to $7\times10^{27}$, which largely reduce the burden of training supernet. As depicted in Figure \ref{supernet_performance}(a), 
with the increase of $l_s$, the amount of search space decreases in a logarithmic tread.

\textbf{Effect of base widths $l_s$ to the performance of supernets.} With BCNetV2, the reduction of the size of search space greatly reduce the burden of training supernet, which can increase the performance of supernet and boost the searched widths. As illustrated in Figure \ref{supernet_performance}(b), during evolutionary, the Top-1 accuracy of searched best performance widths benefit from the increase of base widths $l_s$.

\textbf{Training from scratch performance of searched widths~\wrt~base widths $l_s$.} To examine the effect of base widths $l_s$ to the performance of searched widths, we implement the search on BCNet and BCNetV2~\wrt~different base widths. As shown in Figure \ref{DB_performance}, when $l_s$ is in a small value, the performance of searched architectures of BCNetV2 benefit a lot from the reduction of search space. However, when $l_s$ close to the half of the grouping number $K=20$, the performance
of searched widths of BCNetV2 have a reverse trend of $l_s$. This is because reduction of useless search space can boost the performance, \ie, $l_s \leq 7$, when $l_s$ become too large, some good performance architectures will be lost with the increase of $l_s$. For the ordinal BCNet, the fair training of supernet will be disrupted when $l_s \geq 0$. Therefore, the performance of BCNet always has a reverse trend of $l_s$.

\subsubsection{Effect of iteratively updating of BCNetV2}

With Eq. \eqref{update_eq7}, BCNetV2 can achieve the same memory usage as baseline methods and also the same expectation loss as in Eq. \eqref{eq8}. In this section, we conduct the experiments with two different supernet training ways in Eq. \eqref{eq7} and Eq. \eqref{update_eq7}. As in Table \ref{tab:memory}, with half of memory usage of the BCNet, the BCNetV2 achieves the almost the same performance as BCNet with 2G ResNet50 and 150M MobileNetV2 on ImageNet.

\begin{table}
	\centering
	\caption{Accuracy performance of BCNetV2~\wrt~different updating methods. }  
	\label{tab:memory}
	\vspace{-2mm}
	\begin{tabular}{c||c|c|c|c}
		\hline
		\multicolumn{5}{c}{MobileNetV2} \\ \hline
		Method & FLOPs & Params & Top-1 & Top-5 \\	
		\hline	
		Eq. \eqref{eq7} & 150M & 2.9M & 70.7\% & 89.3\% \\ 
		Eq. \eqref{update_eq7} & 150M & 2.9M & 70.7\% & 89.3\% \\
		AutoSlim \cite{autoslim} & 150M & 3.7M & 69.6\% & 89.2\% \\
		\hline\hline
        \multicolumn{5}{c}{ResNet} \\ \hline
		Method & FLOPs & Params & Top-1 & Top-5 \\	
		\hline	
		Eq. \eqref{eq7} & 2.0G & 17.8M  & 77.3\% & 93.4\%  \\ 
		Eq. \eqref{update_eq7} & 2.0G & 17.7M & 77.3\% & 93.3\% \\ 
		AutoSlim \cite{autoslim} & 2.0G & 20.6M & 75.6\% & 92.7\% \\ \hline\hline
	\end{tabular}
	\vspace{-5mm}
\end{table}

\subsubsection{Effect of prior initial population sampling.}

\begin{figure}[t]
	\centering
	\includegraphics[width=0.68\linewidth,height=0.5\linewidth]{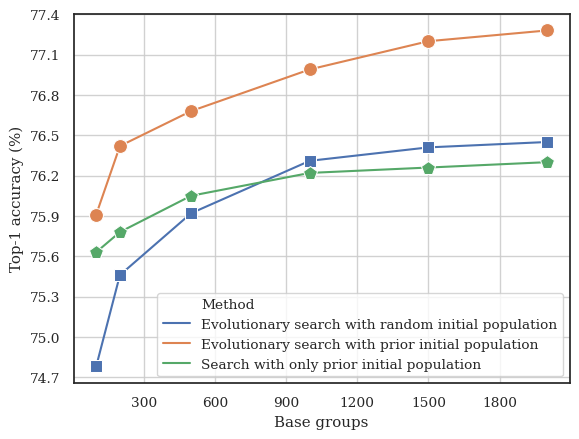}
	\vspace{-3mm}
	\caption{Top-1 accuracy of 0.5 $\times$ ResNet50 on ImageNet dataset by different search methods with the increasing of search numbers. }
	\label{supp_prior}
	\vspace{-4mm}
\end{figure}

Our proposed prior initial population sampling (PIPS) method aims to provide a better initial population for evolutionary search, and the searched optimal width will have higher performance accordingly. Now we want to investigate how the effect of directly leveraging PIPS to search for optimal width. With this aim, we pick up the optimal width with the highest validation Top-1 accuracy after $\{100, 200, 500, 1000, 1500, 2000\}$ of search number of widths, respectively. Then we train them from scratch and report their Top-1 accuracy in Figure \ref{supp_prior}. The search is implemented on ResNet50 on ImageNet dataset with 3 different settings and 0.5$\times$ FLOPs budget, \ie, evolutionary search with random initial population, evolutionary search with prior initial population, and search with the only prior initial population.
\begin{figure*}[t]
	\centering
	\includegraphics[width=\linewidth,height=0.25\linewidth]{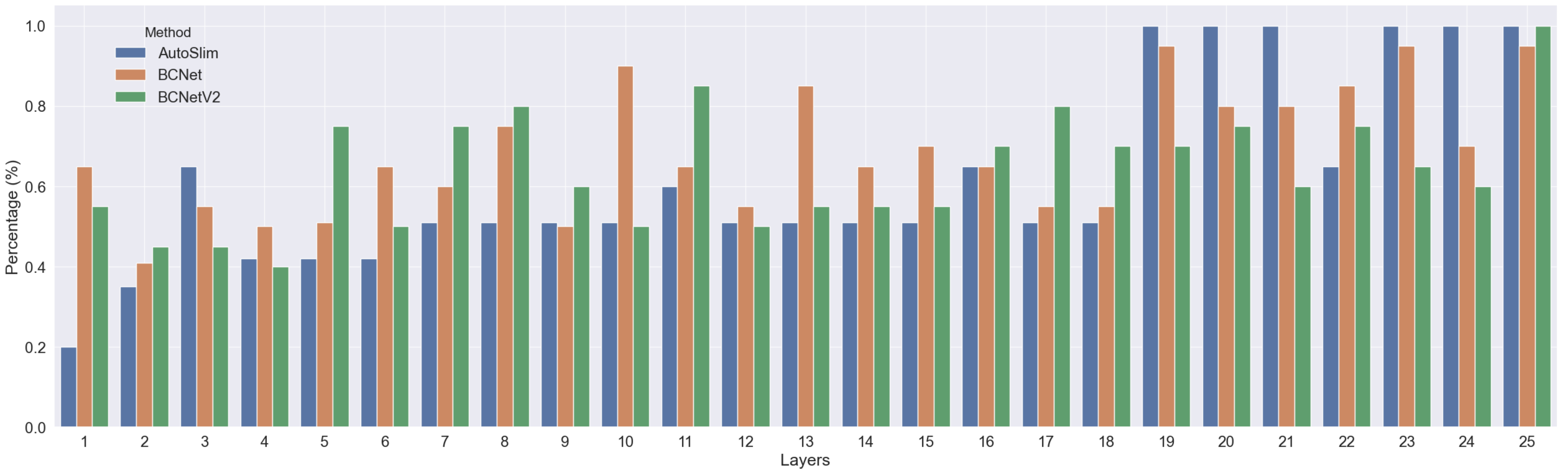}
	\vspace{-5mm}
	\caption{Visualization of searched MobileNetV2 with 305M FLOPs on ImageNet dataset. Both networks are searched on 1.5$\times$ search space as AutoSlim \cite{autoslim}.}
	\label{Compare_AutoSlim}
	\vspace{-5mm}
\end{figure*}
From Figure \ref{supp_prior}, we can know that widths provided by our prior initial population sampling method can surpass those from the random initial population by a larger gap on Top-1 accuracy. Besides, it also should be noticed that evolutionary search benefits more from the increase of search numbers, which indicates that evolutionary search can better utilize the searched width to achieve the optimal result. In addition, with our prior initial population, evolutionary search can get better network width with higher performance (\ie, \orange{orange} line in Figure \ref{supp_prior}), which means that our prior initial population sampling method can provide good initialization for evolutionary algorithm.  Moreover, the Top-1 accuracy of searched models rises slowly after 1000 search numbers, which may imply that the evolutionary algorithm can already find a good solution in this case.

\subsubsection{Transferability of the searched width to object detection task.}
For object detection tasks, a pretrained model on ImageNet dataset is usually leveraged as its backbone. As a result, We take the searched 2G FLOPs ResNet50 as the backbone in detection to examine the transferability of BCNet for other tasks \cite{fpn,frcnn}. The results are reported in Table \ref{detection} for both Faster R-CNN with FPN and RetinaNet, indicating the backbones obtained by BCNet(0.5$\times$) can achieve higher performance to the uniform baseline(0.5$\times$)
\begin{table}[t]
	\centering
\caption{Detection performance with ResNet50 as the backbone.}
\vspace{-2mm}
\label{detection}
\begin{tabular}{l|c|c|c}
		\hline
		Framework&Original (4G)&BCNet (2G)&Uniform(2G)  \\ \hline
		RetinaNet  & 36.4\% & 35.4\% & 34.3\% \\
		Faster R-CNN & 37.3\% & 36.3\% & 35.4\% \\ \hline
\end{tabular}
\vspace{-4mm}
\end{table}

\subsection{Comparison of BCNet, BCNetV2, and AutoSlim \cite{autoslim} under 305M FLOPs}
To intuitively check the effect of BCNet and BCNetV2 with another baseline method, we visualize the network width searched by BCNet, BCNetV2 and the released structure of AutoSlim \cite{autoslim} for 305M-FLOPs MobileNetV2 in Figure \ref{Compare_AutoSlim}. In detail, compared to AutoSlim and BCNet, BCNetV2 saves
more layer widths in the first few layers, and prunes a bit more widths in the last few layers, which is more evenly than AutoSlim and BCNet.

To promote the fair comparison of BCNet, BCNetV2 and AutoSlim, we retrain the released structure of AutoSlim (\ie, 305 FLOPs MobileNetV2) with the same training recipe of ours. Note that we do not include KD for a more fair comparison of AutoSlim \cite{autoslim}, as shown in Table \ref{Experiments_supp_autoslim}.
\begin{table}[t]
	\centering
	\small
	\caption{Performance comparison with AutoSlim \cite{autoslim} of 305M MobileNetV2 on ImageNet by the same training recipe.}
	\label{Experiments_supp_autoslim}
	\vspace{-2mm}
	{\begin{tabular}{c|c|cc|cc}
			\hline
			Methods&FLOPs&Parameters&Top-1&Top-5 \\ \hline
			AutoSlim&305M&5.8M&73.1\%&91.1\% \\
			BCNet&305M&4.8M&73.9\%&92.2\% \\ 
			BCNetV2&305M&4.9M&74.2\%&92.5\% \\ \hline
	\end{tabular}}	
	\vspace{-5mm}
\end{table}

From Table \ref{Experiments_supp_autoslim} and Figure \ref{Compare_AutoSlim}, we can know that BCNet and BCNetV2 retain more widths closer to the input layer, and thus the parameters of our searched structures are lesser than AutoSlim. Moreover, with the same training recipe, BCNet and BCNetV2 achieve  0.8\% and 1.1\% higher on Top-1 accuracy than AutoSlim with 305M FLOPs MobileNetV2, respectively, which indicates the effectiveness of our method.

\section{Visualization and Interpretation of Results}
For intuitively understanding, we visualize our searched networks with various FLOPs in Figure \ref{visualization}. Moreover, for clarity we show the retained ratio of layer widths compared to that of the original models. Note that for MobileNetV2, ResNet50, EfficientNet-B0 and ProxylessNAS with skipping or depthwise layers, we merge these layers which are required to have the same width. 

From Figure \ref{visualization}, we can see that on the whole, with decreasing FLOPs, layer width nearer the input tends to be reduced. However, the last layer is more likely to be retained. This might result from that the last layer is more sensitive to the classification performance, thus it is safely kept when the FLOPs is reduced. In the sequel, we will illustrate more elaborate observations \wrt~each network, and present some intuitions accordingly. 

\subsection{ResNet50 on ImageNet.} We found that when the network is pruned with a large FLOPs budget (\eg, 3G or 2G), width of the first 1$\times$1 convolutional layer (\eg, $2$nd and $5$th layer in Figure \ref{visualization}) of each block in ResNet50 is preferentially reduced, which means 1$\times$1 convolution may contribute less to classification performance. However, when FLOPs drops to a fairly small value (\eg, 570M), channel number of 3$\times$3 convolution (\eg, $3$rd and $6$th layer in Figure \ref{visualization}) will decrease dramatically while that of 1$\times$1 convolution increases instead. This implies that the network will be forced to use more 1$\times$1 convolutions instead of 3$\times$3 convolutions to extract information from feature maps. In addition, this observation also indicates that evolutionary algorithm is more effective than greedy search \wrt~small FLOPs since evolutionary algorithm can always maintain the original search space. Nevertheless, greedy algorithm will greedily prune out more 1$\times$1 convolutions at the beginning, which cannot be recovered for small FLOPs budget.

\subsection{MobileNetV2 on ImageNet and CIFAR-10.} Different from ResNet50, widths of MobileNetV2 decrease more evenly with the reduction of FLOPs. This may be due to the limitation of depthwise convolutions, which requires the output channel number of first 1$\times$1 convolution and the second 3$\times$3 convolution to be the same in MobileNetV2 blocks. Compared from pruning on ImageNet, widths closer to the input layer are more easily to be clipped on CIFAR-10 dataset. This may be because the input of CIFAR-10 is 32$\times$32, which do not need as many widths as ImageNet in the last layer. In addition, when FLOPs is reduced to a fairly small value (\eg, 28M, 44M, and 50M for MobileNetV2), unlike pruning on ImageNet, the width of the last layer of MobileNetV2 on CIFAR-10  decreases rapidly. The reason for this phenomenon may be that MobileNetV2 on ImageNet is forced to classify 1000 categories, while it only needs to deal with 10-way classification on CIFAR-10. Then the width of the last layer on ImageNet tends to be retained, but gets decreased rapidly on CIFAR-10.

\subsection{EfficientNet-B0 on ImageNet.} EfficientNet-B0 shares similar block structure with MobileNetV2. However, the width of EfficientNet-B0 varies more evenly than MobileNetV2, which may be due to its width setting is more better since it is determined by NAS. In detail, compared to the original setting of EfficientNet-B0, for the searched 1$\times$ FLOPs network, the channels of adjacent blocks show opposite fluctuations (\eg, channels of 1,3,5 blocks increase while channels in 2,4,6 blocks decrease). This may mean that the fluctuations of widths are conducive to the performance of searched network structure.

\section{Conclusion}

In this paper, we introduced a new supernet called BCNet to address the training unfairness and corresponding evaluation bias for searching the optimal network width. In our BCNet, each channel is fairly trained and responsible for the same amount of widths. Besides, we propose to reduce the redundant search space and present the BCNetV2 to ensure rigorous training fairness over channels. In addition, we leveraged a stochastic complementary strategy for the training of the BCNet and proposed a prior initial population sampling method to boost the evolutionary search. Concretely,  we also propose the first open-source width benchmark on macro structures named Channel-Bench-Macro for the better comparison of width search algorithms. Extensive experiments have been implemented on ImageNet and CIFAR-10 benchmark datasets to show the superiority of our proposed method to other state-of-the-art channel pruning/network width search methods.


\vspace{-4mm}
\bibliographystyle{IEEEbib}
\bibliography{refs}

\begin{IEEEbiography}[{\includegraphics[width=1in,height=1.25in,clip,keepaspectratio]{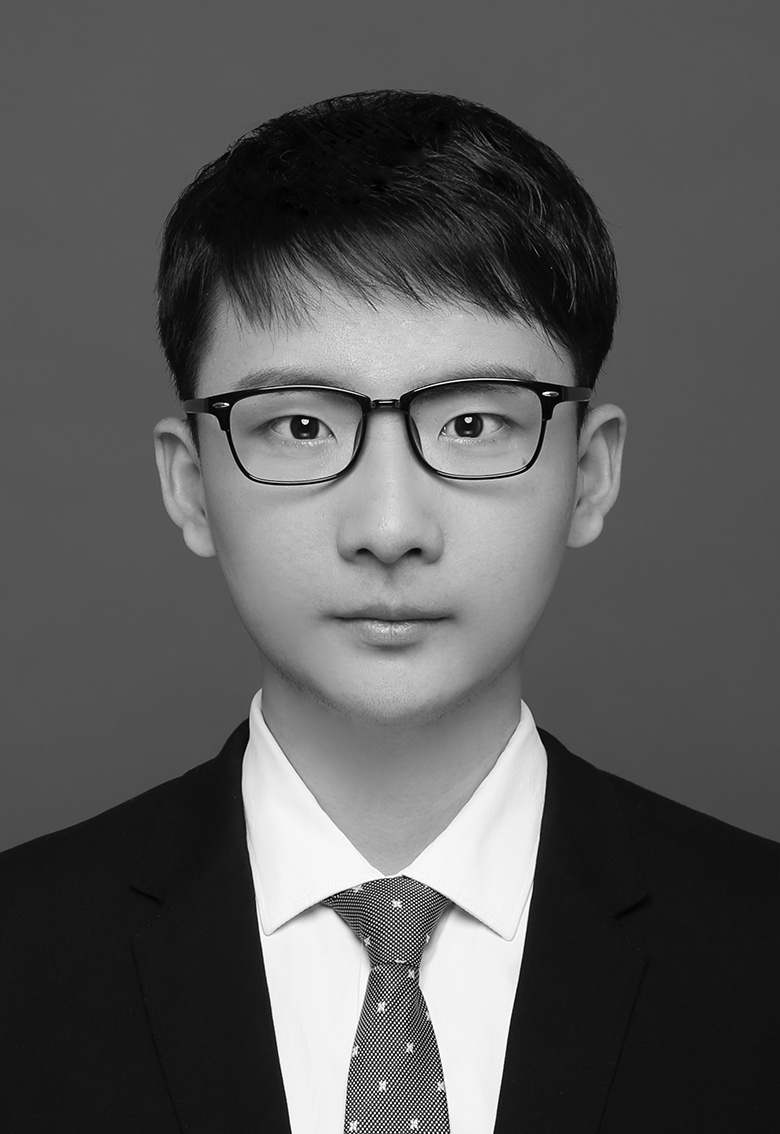}}]{Xiu Su} received his BSc and MSc degree from the Tianjin University. He is currently pursuing the Ph.D. degree in computer science from the University of Sydney. His research interests include pattern recognition and machine learning fundamentals with a focus on neural architecture search, channel number search, detection, transformer.
\end{IEEEbiography}
\begin{IEEEbiography}[{\includegraphics[width=1in,height=1.25in,clip,keepaspectratio]{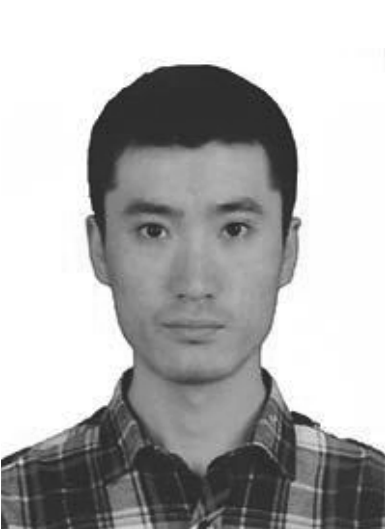}}]{Shan You} is currently a Senior Researcher at SenseTime, and also a post doc at Tsinghua University. Before that, he received a Bachelor of mathematics and applied mathematics (elite class) from Xi'an Jiaotong University, and a Ph.D. degree of computer science from Peking University. His research interests include fundamental algorithms for machine learning and computer vision, such as AutoML, representation learning, light detector and face analysis. He has published his research outcomes in many top tier conferences and transactions.
\end{IEEEbiography}
\begin{IEEEbiography}[{\includegraphics[width=1in,height=1.25in,clip,keepaspectratio]{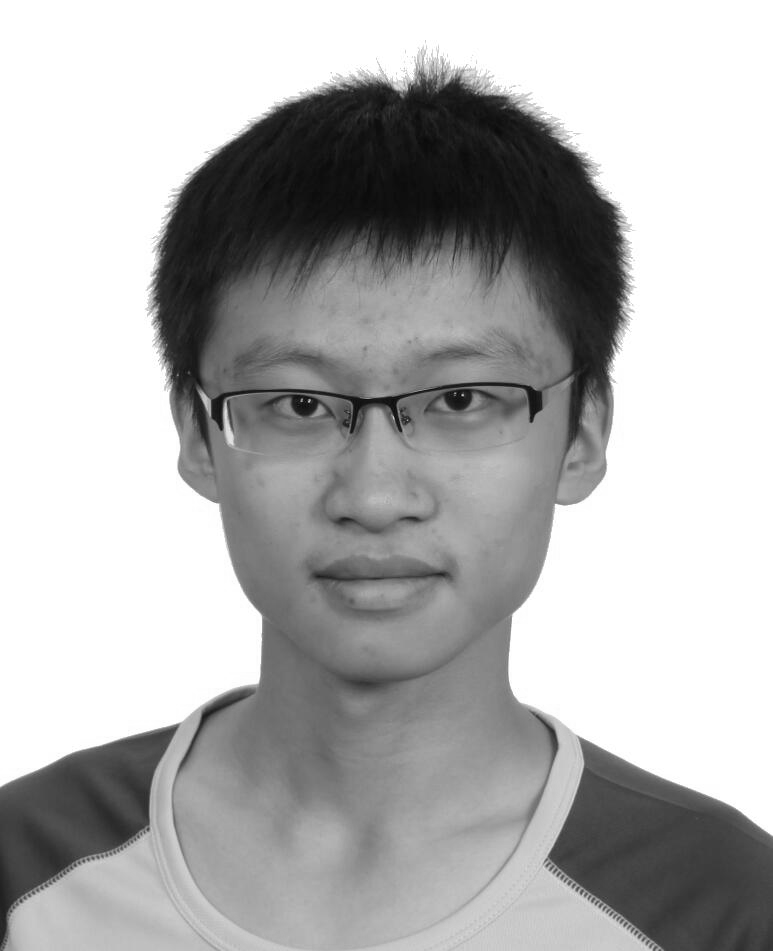}}]{Jiyang Xie} received his B.E. degree in information engineering from Beijing University of Posts and Telecommunications (BUPT), China, in $2017$, where he is currently pursuing the Ph.D. degree. His research interests include pattern recognition and machine learning fundamentals with a focus on applications in image processing, data mining, and deep learning.
\end{IEEEbiography}
\begin{IEEEbiography}[{\includegraphics[width=1in,height=1.25in,clip,keepaspectratio]{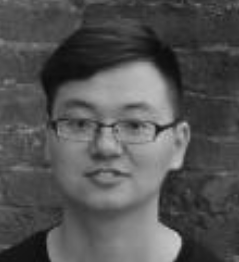}}]{Fei Wang} Fei Wang is the Director of SenseTime Intelligent Automotive Group. He is the head of SenseAuto-Parking engineering and SenseAuto-Cabin research. He leads a vibrant team of 60+ people to develop comprehensive solutions for the intelligent vehicle and deliver 20+ mass production of SenseAuto-Cabin projects in the last 6 years. He has published 20+ papers at CVPR/NIPS/ICCV during the last few years. Fei obtained his Bachelor’s degree and Master's degree from Beijing University of Posts and Telecommunications. Currently, he is a Ph.D. student at the University of Science and Technology of China. His research interests include Automotive Drive System, AI Chip, Deep Learning, etc.
\end{IEEEbiography}
\begin{IEEEbiography}[{\includegraphics[width=1in,height=1.25in,clip,keepaspectratio]{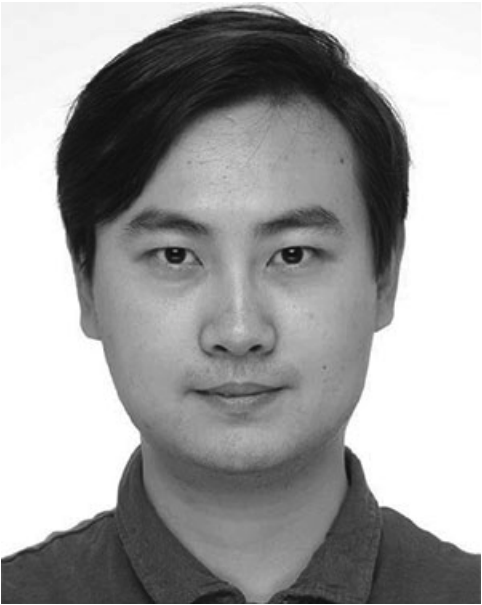}}]{Chen Qian} received the BEng degree from the Institute for Interdisciplinary Information Science, Tsinghua University, in 2012, and the MPhil degree from the Department of Information Engineering, the Chinese University of Hong Kong, in 2014. He is currently working at SenseTime as research director. His research interests include human-related computer vision and machine learning problems.
\end{IEEEbiography}
\begin{IEEEbiography}[{\includegraphics[width=1in,height=1.25in,clip,keepaspectratio]{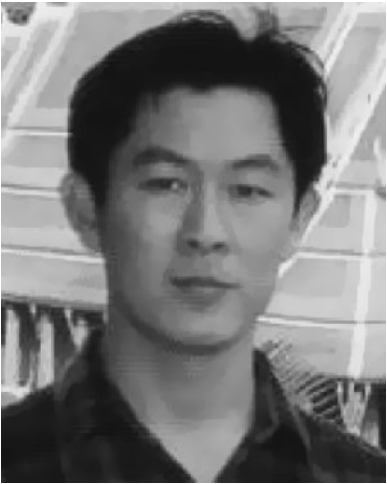}}]{Changshui Zhang} received the BSc degree in mathematics from Peking University, Beijing, China, in 1986, and the PhD degree from Tsinghua University, Beijing, China, in 1992. In 1992, he joined the Department of Automation, Tsinghua University, where he is currently a professor. His interests include pattern recognition, machine learning, etc. He has authored more than 200 papers. He currently serves on the editorial board of the journal Pattern Recognition. He is a member of the IEEE.
\end{IEEEbiography}
\begin{IEEEbiography}[{\includegraphics[width=1in,height=1.25in,clip,keepaspectratio]{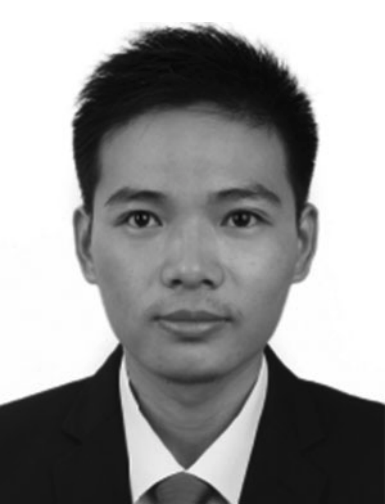}}]{Chang Xu} is Senior Lecturer and ARC DECRA Fellow at the School of Computer Science, University of Sydney. He received the Ph.D. degree from Peking University, China. His research interests lie in machine learning algorithms and related applications in computer vision. He has published over 90 papers in prestigious journals and top tier conferences. He has received several paper awards, including Distinguished Paper Award in IJCAI 2018. He regularly severed as the PC member or senior PC member for many conferences, e.g. NeurIPS, ICML, ICLR, CVPR, ICCV, IJCAI and AAAI. He has been recognized as Top Ten Distinguished Senior PC Member in IJCAI 2017.
\end{IEEEbiography}
\end{document}